\documentclass[10pt,twocolumn]{IEEEtran}

\usepackage{graphicx}
\usepackage{epsfig}

\usepackage{url}

\usepackage{amsmath}

\usepackage{amssymb}
\usepackage{epsfig}
\usepackage{epstopdf}

\usepackage{array}

\begin{document}
%
\title{Compressed Sensing for Energy-Efficient Wireless Telemonitoring of Noninvasive Fetal ECG via Block Sparse Bayesian Learning}

\author{Zhilin Zhang$^*$, \IEEEmembership{Student Member, IEEE}, Tzyy-Ping Jung, \IEEEmembership{Senior Member, IEEE} \\Scott Makeig, \IEEEmembership{Member, IEEE}, Bhaskar D. Rao, \IEEEmembership{Fellow, IEEE}
\thanks{The work was supported by NSF grants CCF-0830612, CCF-1144258 and DGE-0333451, and was in part supported by Army Research Lab, Army Research Office, Office of Naval Research, and Defense Advanced Research Projects Agency (DARPA). \emph{Asterisk indicates corresponding author}.}
\thanks{Z.Zhang and B.D.Rao are with the Department of Electrical and Computer Engineering, University of California, San Diego, La Jolla, CA 92093-0407, USA. Email: zhangzlacademy@gmail.com (Z.Z.), brao@ucsd.edu (B.D.R.)}
\thanks{T.-P. Jung and S. Makeig are with the Swartz Center for Computational Neuroscience, University of California, San Diego, La Jolla, CA 92093-0559, USA. Email:\{jung,scott\}@sccn.ucsd.edu}
}

\markboth{Published in IEEE Transactions on Biomedical Engineering, vol. 60, no. 2, pp. 300-309, 2013}{Zhang \MakeLowercase{\textit{et al.}}: }

\maketitle

\begin{abstract}
Fetal ECG (FECG) telemonitoring is an important branch in telemedicine. The design of a telemonitoring system via a wireless body-area network with low energy consumption for ambulatory use is highly desirable. As an emerging technique, compressed sensing (CS) shows great promise in compressing/reconstructing data with low energy consumption. However, due to some specific characteristics of raw FECG recordings such as non-sparsity and strong noise contamination, current CS algorithms generally fail in this application.

This work proposes to use the block sparse Bayesian learning (BSBL) framework to compress/reconstruct non-sparse raw FECG recordings. Experimental results show that the framework can reconstruct the raw recordings with high quality. Especially, the reconstruction does not destroy the interdependence relation among the multichannel recordings. This ensures that the independent component analysis decomposition of the reconstructed recordings has high fidelity. Furthermore, the framework allows the use of a sparse binary sensing matrix with much fewer nonzero entries to compress recordings. Particularly, each column of the matrix can contain only \emph{two} nonzero entries. This shows the framework, compared to other algorithms such as current CS algorithms and wavelet algorithms, can greatly reduce code execution in CPU in the data compression stage.
\end{abstract}


\IEEEpeerreviewmaketitle

\section{Introduction}
\label{sec:intro}

Noninvasive monitoring of fetal ECG (FECG) is an important approach to monitor the health of fetuses. The characteristic parameters of an FECG, such as heart beat rates, morphology, and dynamic behaviors, can be used for diagnosis of fetal development and disease. Among these parameters, the heart beat rate is the main index of fetal assessment for high-risk pregnancies \cite{smith2008fetal}. For example, abnormal patterns (decelerations, loss of high-frequency variability, and pseudo-sinusoidal) of fetal heart beat rates are generally indicative of fetal asphyxia \cite{hasan2009detection}.

However,  noninvasive acquisition of clean FECGs from maternal abdominal recordings is not an easy problem. This is because FECGs are very weak, and often embedded in strong noise and interference, such as maternal ECGs (MECGs), instrumental noise, and artifacts caused by muscles. Further, the gestational age and the position of fetuses also affect the strength of FECGs. Up to now various signal processing and machine learning methods have been proposed to obtain FECGs, such as adaptive filtering \cite{sameni2007nonlinear}, wavelet analysis \cite{addison2005wavelet}, and blind source separation (BSS)/independent component analysis (ICA) \cite{de2000fetal}. For example, the problem of extracting clean FECGs from raw FECG recordings can be well modeled as an instantaneous ICA mixture model, in which the raw recordings are viewed as the linear mixture of a number of independent (or uncorrelated) sources including FECG components, MECG components, and various noise components \cite{de2000fetal}. Interested readers can refer to \cite{hasan2009detection,addison2005wavelet,sameni2010review} for good surveys on these techniques.

Traditionally, pregnant women are required to frequently visit hospitals to get resting FECG monitoring. Now, the trend and desire is to allow pregnant women to receive ambulatory monitoring of FECGs. For example, pregnant women can stay at home, where FECGs are collected through wireless telemonitoring. In such a telemonitoring system, a wireless body-area network (WBAN) \cite{specialissue} integrates a number of sensors attached on a patient's skin, and uses ultra-low-power short-haul radios (e.g., Bluetooth) in conjunction with nearby smart-phones or handheld devices to communicate via the Internet with the health care provider in a remote terminal. Telemonitoring is a convenient way for patients to avoid frequent hospital visits and save lots of time and medical expenses.

Among many constraints in WBAN-based telemonitoring systems \cite{cao2009enabling}, the energy consumption is a primary design constraint \cite{milenkovic2006wireless}.  It is necessary to reduce energy consumption as much as possible, since a WBAN is often battery-operated. This has to be done in several ways. One way is that on-sensor computation should be minimum. Another is that data should be compressed before transmission (the compressed data will be used to reconstruct the original data in remote terminals). Unfortunately, most conventional data compression techniques such as wavelet-based algorithms dissipate lots of energy \cite{mamaghanian2011compressed}. Therefore, new compression techniques are needed urgently.

Compressed sensing (CS) \cite{donoho2006compressed}, an emerging signal processing technique, is a promising tool to cater to the two constraints. It uses a simple linear transform (i.e., a sensing matrix) to compress a signal, and then reconstructs it by exploiting its sparsity. The sparsity refers to the characteristics that most entries of the signal are zero. When CS is used in WBAN-based telemonitoring systems, the compression stage is completed on data acquisition module before transmission, while the reconstruction stage is completed on workstations/computers at remote receiving terminals. Based on a real-time ECG telemonitoring system, Mamaghanian et al. \cite{mamaghanian2011compressed} showed that when using a sparse binary matrix as the sensing matrix, CS can greatly extend sensor lifetime and reduce energy consumption while achieving competitive compression ratio, compared to a wavelet-based compression method. They also pointed out that when the data collection and the compression are implemented together by analog devices before analog-to-digital converter (ADC), the energy consumption can be further reduced.

Although CS has achieved some successes in adult ECG telemonitoring \cite{mamaghanian2011compressed,eduardo2010implementation,Dixon2012}, it encounters difficulties in FECG telemonitoring. These difficulties essentially come from the conflict between more strict energy constraint in FECG telemonitoring systems and non-sparsity of raw FECG recordings.

The energy constraint is more strict in FECG telemonitoring systems due to the large number of sensors deployed. Generally, the number of sensors to receive raw FECG recordings ranges from 8 to 16, and sometimes extra sensors are needed to record maternal physiological signals (e.g., blood pressure, MECG, and temperature). The large number of sensors indicates large energy dissipated in on-sensor computation. Given limited energy, this requires the systems to perform as little on-sensor computation as possible. For example, filtering before data compression may be prohibited. For CS algorithms, this means that they are required to directly compress raw FECG recordings with none or minimum pre-processing.

However, raw FECG recordings are non-sparse, which seriously deteriorates reconstruction quality of CS algorithms. Raw FECG recordings differ from adult ECG recordings
in that they are unavoidably contaminated by a number of strong noise and interference, as discussed previously. Most CS algorithms have
difficulty in directly reconstructing such non-sparse signals. Although some strategies have been proposed to deal with non-sparse signals, they may not be helpful in this application.

This study proposes to use the Block Sparse Bayesian Learning (BSBL) framework \cite{zhang2012TSP,zhang2011IEEE}
to address these challenges. The BSBL framework, as a new framework solving the CS problems, has some advantages
over conventional CS frameworks. It provides large flexibility to exploit spatial \cite{zhang2012TSP}, temporal \cite{zhang2011IEEE}, and dynamic structure \cite{zhang2011ICML} of signals. Algorithms derived from this framework have superior performance compared to most existing CS algorithms. Here, we present two interesting properties of the BSBL framework. One is its ability to reconstruct non-sparse signals with high quality. The other is its ability to exploit unknown structure of signals for better reconstruction quality. These two properties make the BSBL framework successful in wireless FECG telemonitoring.

The rest of the paper is organized as follows. Section \ref{sec:bsbl} introduces basic CS models and the BSBL framework. Section \ref{sec:experiments} shows experimental results on two typical FECG datasets.  Section \ref{sec:issues}  studies some issues related to the BSBL framework and our experiments. Discussions and conclusions are given in the last two sections.

\section{Compressed Sensing and Block Sparse Bayesian Learning}
\label{sec:bsbl}

As in the `\emph{digital CS}' paradigm in \cite{mamaghanian2011compressed}, we assume signals have passed through the analog-to-digital converter (ADC).

\subsection{Compressed Sensing and Associated Models}
\label{subsec:cs}

CS is a new signal compression paradigm which relies on the sparsity of signals to compress and reconstruct. The \emph{basic noisy model} can be expressed as
\begin{eqnarray}
\mathbf{y}= \mathbf{\Phi} \mathbf{x} + \mathbf{v},
\label{equ:SMV model}
\end{eqnarray}
where $\mathbf{x}\in \mathbb{R}^{N \times 1}$ is the signal to compress/reconstruct with length $N$. $\mathbf{\Phi} \in \mathbb{R}^{M \times N} (M \ll N)$ is a designed sensing matrix which linearly compresses  $\mathbf{x}$. Any $M$ columns of $\mathbf{\Phi}$ are linearly independent. $\mathbf{v}$ is a noise vector modeling errors incurred during this compression procedure or noise in the CS system. In our application $\mathbf{x}$ is a segment from a raw FECG recording, $\mathbf{y}$ is the compressed data which will be transmitted via a WBAN to a remote terminal, and $\mathbf{v}$ can be ignored. Thus the model used in our application is a noiseless model, expressed as
\begin{eqnarray}
\mathbf{y}= \mathbf{\Phi} \mathbf{x}.
\label{equ:SMV model_noiseless}
\end{eqnarray}
In the remote terminal, using the designed sensing matrix $\mathbf{\Phi}$, a CS algorithm reconstructs $\mathbf{x}$ from the compressed data $\mathbf{y}$. Note that reconstructing $\mathbf{x}$ is an underdetermined inverse problem. By exploiting the sparsity of $\mathbf{x}$ it is possible to exactly recover it in noiseless cases, or recover it with small errors in the presence of noise.

In many applications the signal $\mathbf{x}$ is not sparse, but sparse in some transformed domains such as the wavelet domain. This means, $\mathbf{x}$ can be expressed as $\mathbf{x} =\mathbf{\Psi} \boldsymbol{\theta}$, where $\mathbf{\Psi} \in \mathbb{R}^{N \times N}$ is an orthonormal basis matrix of a transformed domain
and $\boldsymbol{\theta}$ is the representation coefficient vector which is sparse. Thus the model (\ref{equ:SMV model_noiseless}) becomes
\begin{eqnarray}
\mathbf{y}= \mathbf{\Phi} \mathbf{\Psi} \boldsymbol{\theta}  = \boldsymbol{\Omega} \boldsymbol{\theta},
\label{equ:CS_model}
\end{eqnarray}
where $\boldsymbol{\Omega} \triangleq \mathbf{\Phi} \mathbf{\Psi}$. Since $\boldsymbol{\theta}$ is sparse, a CS algorithm can first reconstruct $\boldsymbol{\theta}$ using $\mathbf{y}$ and $\boldsymbol{\Omega}$, and then reconstruct $\mathbf{x}$ by $\mathbf{x} =\mathbf{\Psi} \boldsymbol{\theta}$. This method is useful for some kinds of signals. But as shown in our experiments later, this method still cannot help existing CS algorithms to reconstruct raw FECG recordings.

Sometimes the signal $\mathbf{x}$ itself contains noise (called `\emph{signal noise}'). That is, $\mathbf{x} = \mathbf{u} + \mathbf{n}$, where $\mathbf{u}$ is the clean signal and $\mathbf{n}$ is the signal noise. Thus the model (\ref{equ:SMV model_noiseless}) becomes
\begin{eqnarray}
\mathbf{y}= \mathbf{\Phi} \mathbf{x} = \mathbf{\Phi} (\mathbf{u} + \mathbf{n}) = \mathbf{\Phi} \mathbf{u} + \mathbf{\Phi} \mathbf{n} = \mathbf{\Phi} \mathbf{u} + \mathbf{w},
\label{equ:noise_folding}
\end{eqnarray}
where $\mathbf{w} \triangleq \mathbf{\Phi} \mathbf{n}$ is a new noise vector. This model can be viewed as a basic noisy CS model.

Most natural signals have rich structure. A widely existing structure in natural signals is block structure \cite{BOMP}. A sparse signal with this structure can be viewed as a concatenation of a number of blocks, i.e
\begin{eqnarray}
\mathbf{x} = [ \underbrace{x_1,\cdots,x_{h_1}}_{\mathbf{x}_1^T},   \cdots,  \underbrace{x_{h_{g-1}+1},\cdots,x_{h_g}}_{\mathbf{x}_g^T}]^T
\label{equ:partition}
\end{eqnarray}
where $\mathbf{x}_i \in \mathbb{R}^{h_i \times 1}$, and $h_i (i=1,\cdots,g)$ are not necessarily identical. Among these blocks, only a few blocks are non-zero. A signal with this block structure is called a \emph{block sparse signal}, and the model (\ref{equ:SMV model}) or (\ref{equ:SMV model_noiseless}) with the block partition (\ref{equ:partition}) is called a \emph{block sparse model}. It has been known that exploiting such block structure can further improve reconstruction performance \cite{zhang2012TSP,ModelCS,BOMP}.

\begin{figure}[tbp]
\centering
\includegraphics[width=8cm,height=5cm]{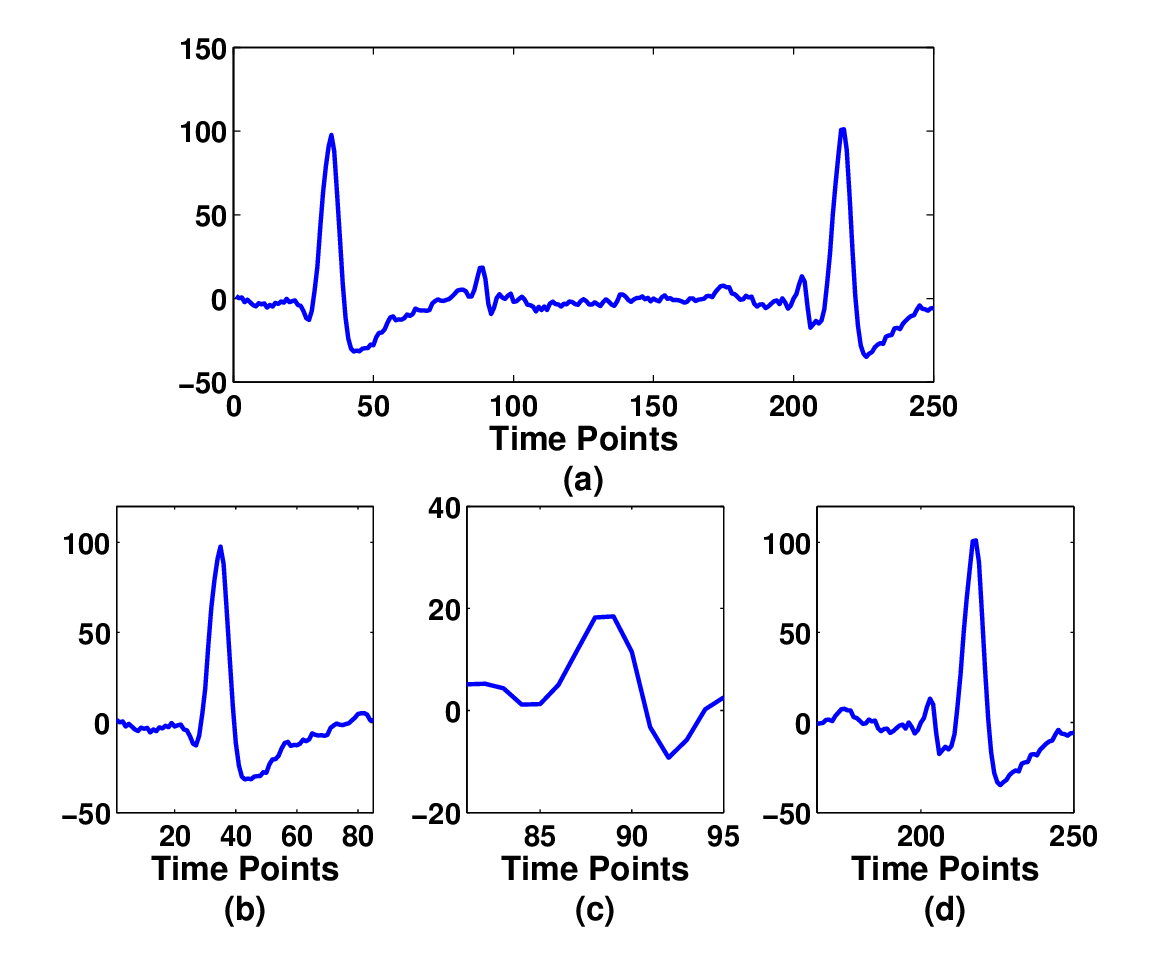}
\caption{Close-up of the second channel recording in the DaISy Dataset in Fig.\ref{fig:ecg_wholedata}. (a) A segment of the first 250 time points of the recording. (b) A sub-segment containing a QRS complex of the MECG. (c) A sub-segment containing a QRS complex of the FECG. (d) A sub-segment showing a QRS complex of the FECG contaminated by a QRS complex of the MECG.}
\label{fig:ecgsegment}
\end{figure}

A raw FECG recording can be roughly viewed as a block sparse signal contaminated by signal noise. Figure \ref{fig:ecgsegment} (a) plots a segment of a raw FECG recording. In this segment the parts from 20 to 60, from 85 to 95, and from 200 to 250 time points can be viewed as three significant non-zero blocks. Other parts can be viewed as concatenations of zero blocks. And the whole segment can be viewed as  a clean signal contaminated by signal noise. Note that although the block partition can be roughly determined by observing the raw recording, it is unknown in practical FECG telemonitoring. Hence, a raw FECG recording can be modeled as a block sparse signal with unknown block partition and unknown signal noise in a noiseless environment.

Reconstructing $\mathbf{x}$ while exploiting its unknown block partition is very difficult. Up to now only several CS algorithms have been proposed for this purpose \cite{Huang2009,bcs_mcmc,faktor2010exploiting}. Recently we proposed the BSBL framework \cite{zhang2012TSP,zhang2011IEEE}, and derived a family of algorithms \cite{zhang2012TSP}. Their ability to reconstruct non-sparse but structured signals endows them with superior performance over  existing algorithms. The following subsection will briefly introduce it. Interested readers can refer to \cite{zhang2012TSP,zhang2011IEEE,zhang2011TR} for details.

\subsection{Block Sparse Bayesian Learning}
\label{subsec:bsbl}

For a block sparse signal $\mathbf{x}$ of the form (\ref{equ:partition}), the BSBL framework models each block $\mathbf{x}_i \in \mathbb{R}^{d_i \times 1}$  as a parameterized multivariate Gaussian distribution:
\begin{eqnarray}
p(\mathbf{x}_i; \gamma_i, \mathbf{B}_i) \sim  \mathcal{N}(\textbf{0},\gamma_i \mathbf{B}_i), \quad i=1,\cdots,g
\label{equ:blockassumption}
\end{eqnarray}
where $\gamma_i$ is a nonnegative parameter controlling the block-sparsity of $\mathbf{x}$. When $\gamma_i=0$, the corresponding $i$-th block, i.e., $\mathbf{x}_i$, becomes a zero block. $\mathbf{B}_i \in \mathbb{R}^{d_i \times d_i}$ is a positive definite matrix capturing the correlation structure of the $i$-th block. With the assumption that blocks are mutually uncorrelated, according to (\ref{equ:blockassumption}), the prior of $\mathbf{x}$ is $p(\mathbf{x}) \sim  \mathcal{N}(\textbf{0},\mathbf{\Sigma}_0)$, where $\mathbf{\Sigma}_0$ is a block-diagonal matrix with the $i$-th principal block given by $\gamma_i \mathbf{B}_i$. The noise vector is assumed to satisfy a multivariate Gaussian distribution, namely $p(\mathbf{v}) \sim  \mathcal{N}(\textbf{0},\lambda \mathbf{I})$, where $\lambda$ is a positive scalar and $\mathbf{I}$ is the identity matrix.

Thus, the estimate of $\mathbf{x}$ can be readily obtained by the Maximum-A-Posterior estimation, provided all the parameters $\{\gamma_i,\mathbf{B}_i\}_1^g$ and $\lambda$ have been estimated, which is usually carried out using the Type-II maximum likelihood estimation \cite{Tipping2001}.

Three iterative algorithms \cite{zhang2012TSP,zhang2012ICASSP} have been derived to reconstruct $\mathbf{x}$. We choose the Bound-Optimization based Block SBL algorithm, denoted by BSBL-BO, to show the ability of BSBL in FECG telemonitoring. Its  reconstruction of  non-sparse signals is achieved by setting a $\gamma_i$-pruning threshold \footnote{Such threshold is used in most SBL algorithms \cite{Tipping2001}, not merely in the BSBL framework.} to a small value. The threshold is used to prune out small $\gamma_i$ during iterations of the algorithm. A smaller value of the threshold means fewer $\gamma_i$ are pruned out, and thus fewer blocks in $\mathbf{x}$ become zero blocks. Consequently, the estimated $\mathbf{x}$ is less sparse. In our experiments we set the threshold to $0$, i.e., disabling the pruning mechanism.

BSBL-BO (and other algorithms derived in \cite{zhang2012TSP}) has another ability, namely exploring and exploiting correlation structure in each block $\mathbf{x}_i$ (i.e., the intra-block correlation) through estimating the matrices $\mathbf{B}_i$. Our experiments will show this ability makes it achieve better reconstruction performance.

Although BSBL-BO needs users to define the block partition (\ref{equ:partition}), it does not require the user-defined block partition to be the same as the true block partition \cite{zhang2012TSP,zhang2012ICASSP}. Later we will further confirm this.

\section{Experiments}
\label{sec:experiments}

Experiments were carried out using two real-world raw FECG datasets \footnote{Experiment codes can be downloaded at \url{http://dsp.ucsd.edu/~zhilin/BSBL.html} or \url{https://sites.google.com/site/researchbyzhang/bsbl}.}. Both datasets are widely used in the FECG community. In the first dataset, the FECG is barely visible, while in the second dataset the FECG is invisible. Thus the two datasets provide a good diversity
of FECG recordings to verify the efficacy of  our algorithm under various situations.

For algorithm comparison, this study chooses ten representative CS algorithms. Each of them represents a family of algorithms and has top-tier performance in its family. Thus, the comparison conclusions could be generalized to other related CS algorithms.

In each experiment, all the CS algorithms used the same sensing matrix to compress FECG recordings. Thus the energy consumption of each CS algorithm was the same \footnote{Reconstruction of FECG recordings is done by software in remote terminals and thus it does not cost energy of WBANs.}. Therefore we only present reconstruction results.

In adult ECG telemonitoring or other applications, reconstruction performance is generally measured by comparing reconstructed recordings with original recordings using the mean square error (MSE) as a performance index. However, in our application reconstructing raw FECG recordings is not the final goal; the reconstructed recordings are further processed to extract a clean FECG by other advanced signal processing techniques such as BSS/ICA and nonlinear filtering. Due to the infidelity of MSE for structured signals \cite{wang2009mean}, it is hard to see how the final FECG extraction is affected by errors in reconstructed recordings measured by MSE. Thus, a more direct measure is to compare the extracted FECG from the reconstructed recordings with the extracted one from the original recordings. This study used BSS/ICA algorithms to extract a clean FECG from reconstructed recordings and a clean FECG from original recordings, and then calculated the Pearson correlation between the two extracted FECGs.

\subsection{The DaISy Dataset}
\label{subsec:data1}

Figure \ref{fig:ecgsegment} shows a segment from the DaISy dataset \cite{ECGdata}. Two QRS complexes of the MECG can be clearly seen from this segment, and two QRS complexes of the FECG can be seen but not very clearly. We can clearly see that the segment is far from sparse; its every entry is non-zero. This brings a difficulty to existing CS algorithms to reconstruct it.

To compress the data we used a randomly generated sparse binary sensing matrix of the size $125\times 250$. Its each column contained 15 entries of 1s, while other entries were zero.

For the BSBL-BO algorithm, we defined its block partition according to (\ref{equ:partition}) with $h_1=\cdots=h_g=25$. Section \ref{sec:issues}  will show that the algorithm is not sensitive to the block partition. The algorithm was employed in two ways. The first way was allowing it to adaptively learn and exploit intra-block correlation. The second way was preventing it from exploiting intra-block correlation, i.e.  by fixing the matrices $\mathbf{B}_i (\forall i)$ to identity matrices.

\begin{figure}[tbp]
\centering
\includegraphics[width=7cm,height=7cm]{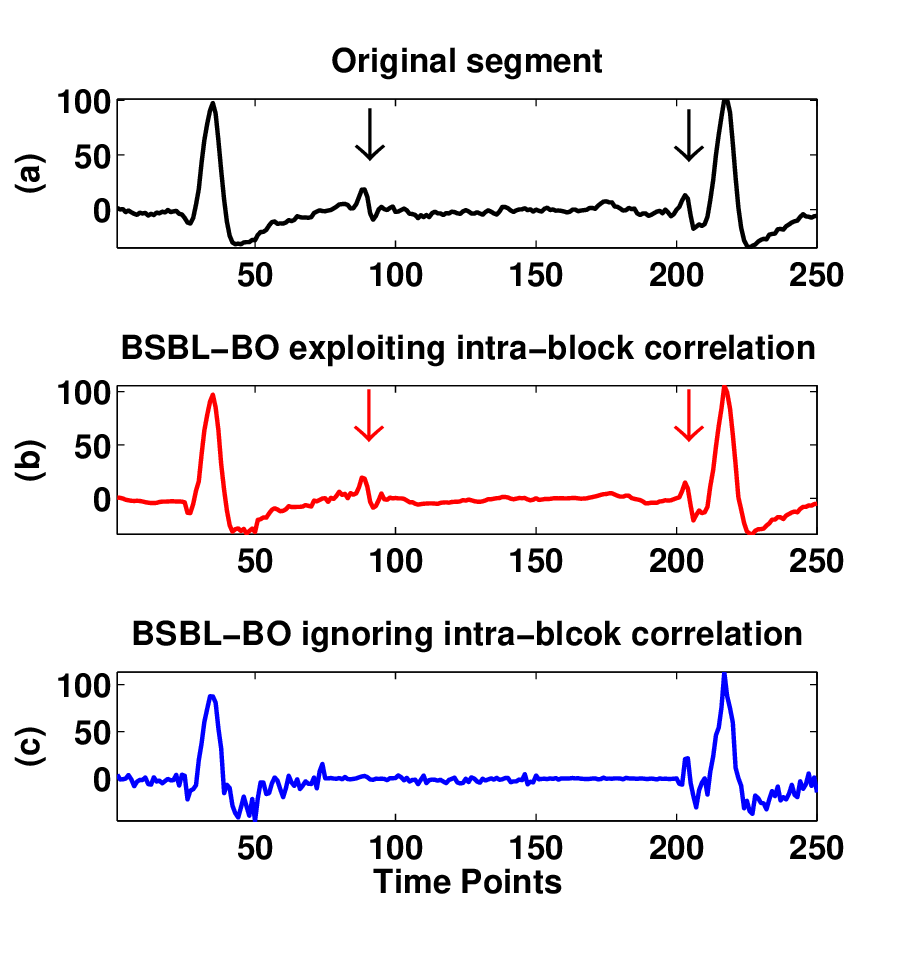}
\caption{Comparison of the original segment and the reconstructed segments by BSBL-BO with and without exploiting intra-block correlation. (a) The original FECG segment. (b) The reconstructed segment by BSBL-BO when exploiting intra-block correlation. (c) The reconstructed segment by BSBL-BO when not exploiting intra-block correlation. The arrows indicate QRS complexes of the FECG. }
\label{fig:exp1_BSBL}
\end{figure}

The results are shown in Figure \ref{fig:exp1_BSBL}, from which we can see that exploiting intra-block correlation allowed the algorithm to reconstruct the segment with high quality. When the correlation was not exploited, the reconstruction quality was very poor; for example, the first QRS complex of the FECG was missing in the reconstructed segment (Figure \ref{fig:exp1_BSBL} (c)).

\begin{figure}[tbp]
\begin{minipage}[b]{.48\linewidth}
  \centering
  \centerline{\epsfig{figure=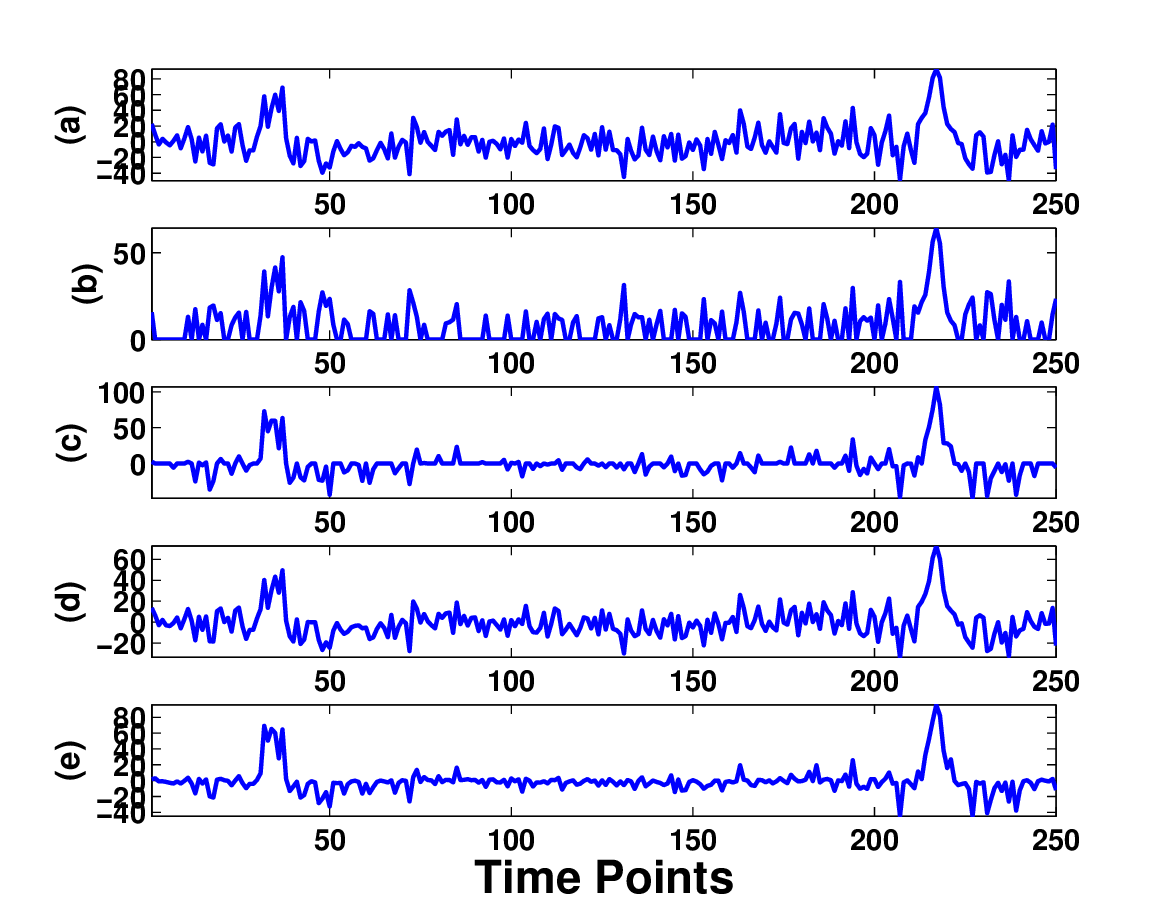,width=4.5cm,height=4.5cm}}
\end{minipage}
\hfill
\begin{minipage}[b]{0.48\linewidth}
  \centering
  \centerline{\epsfig{figure=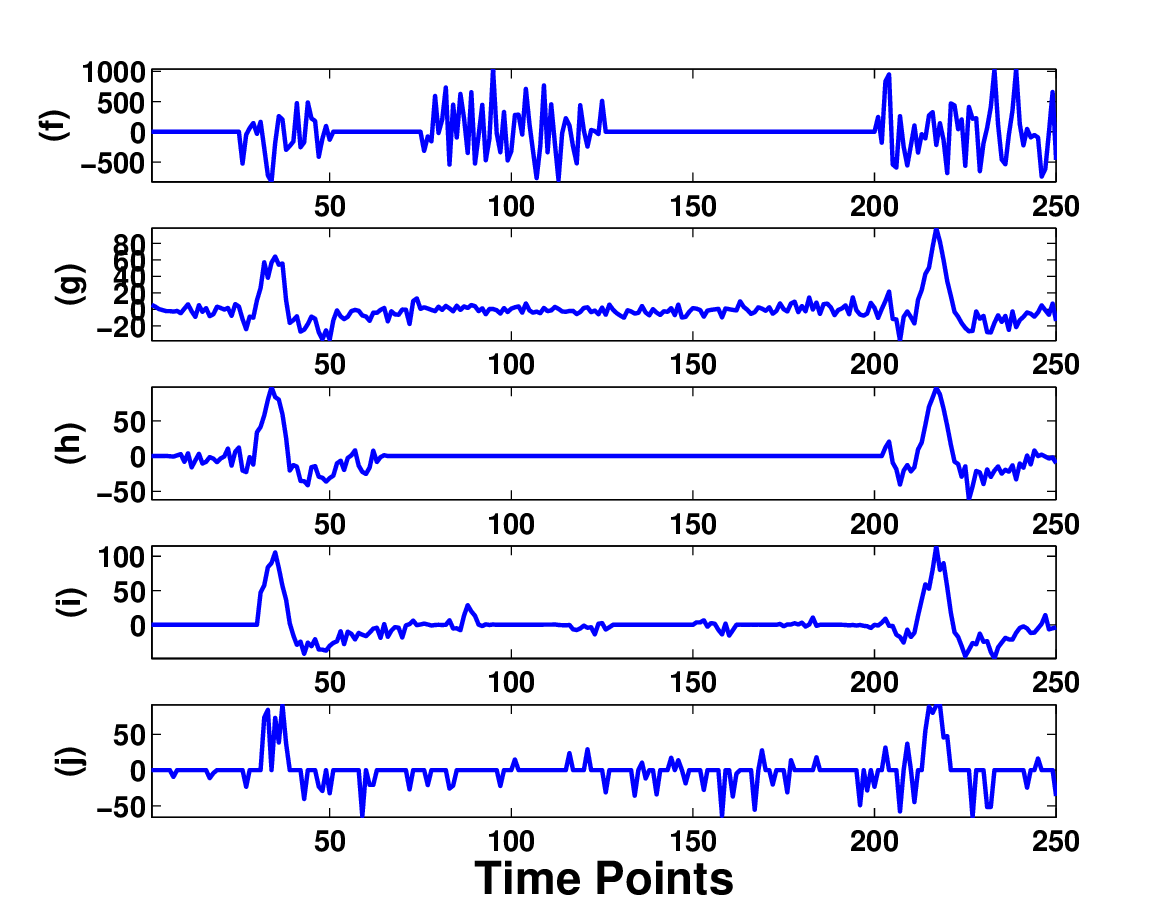,width=4.5cm,height=4.5cm}}
\end{minipage}
\caption{Recovery results of compared algorithms. From (a) to (j), they are the results by (a) Elastic Net, (b) CoSaMP, (c) Basis Pursuit, (d) SL0, (e) EM-GM-AMP, (f) Block-OMP, (g) Block Basis Pursuit, (h) CluSS-MCMC, (i) StructOMP, and (j) BM-MAP-OMP, respectively.}
\label{fig:exp1_CS}
\end{figure}

\begin{figure}[tbp]
\begin{minipage}[b]{.48\linewidth}
  \centering
  \centerline{\epsfig{figure=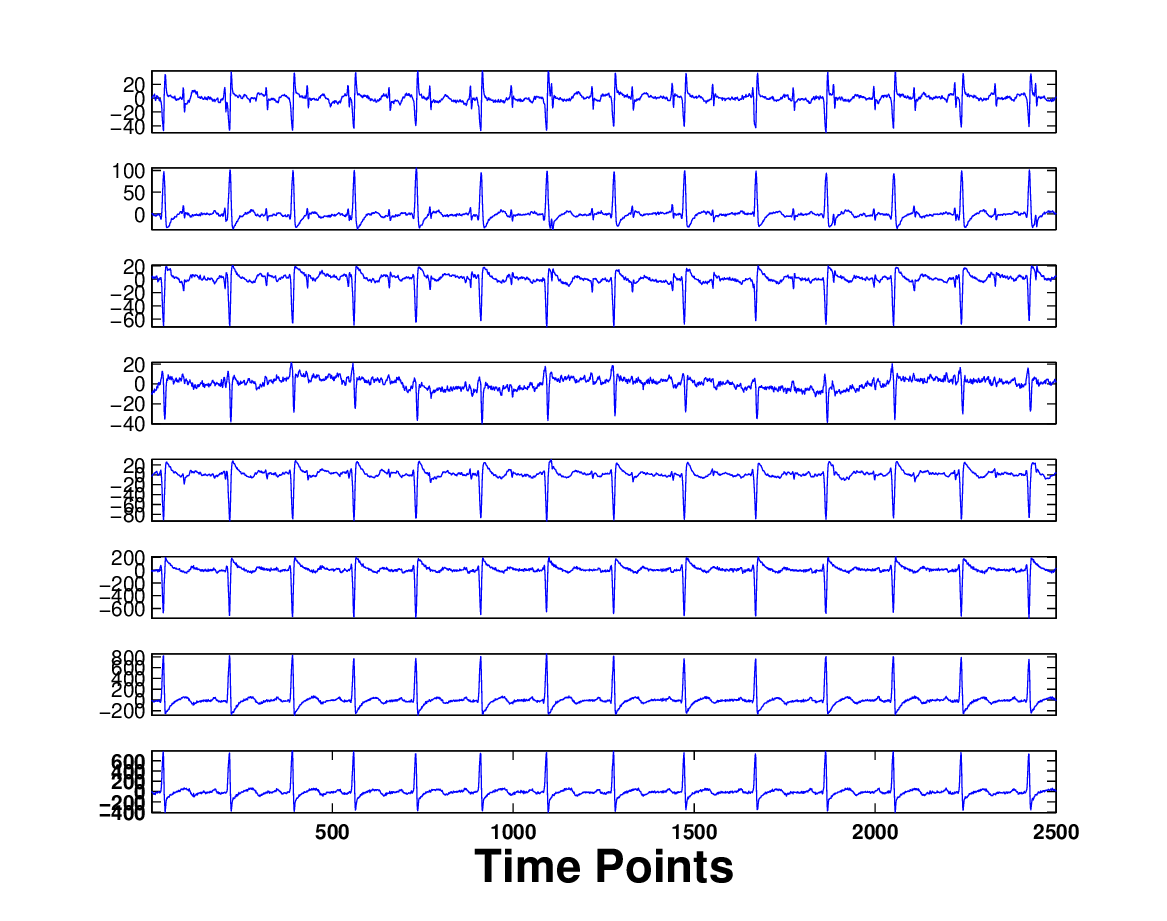,width=4.8cm,height=5cm}}
  \centerline{\footnotesize{(a) Original DaISy Dataset}} \hfill
\end{minipage}
\begin{minipage}[b]{0.48\linewidth}
  \centering
  \centerline{\epsfig{figure=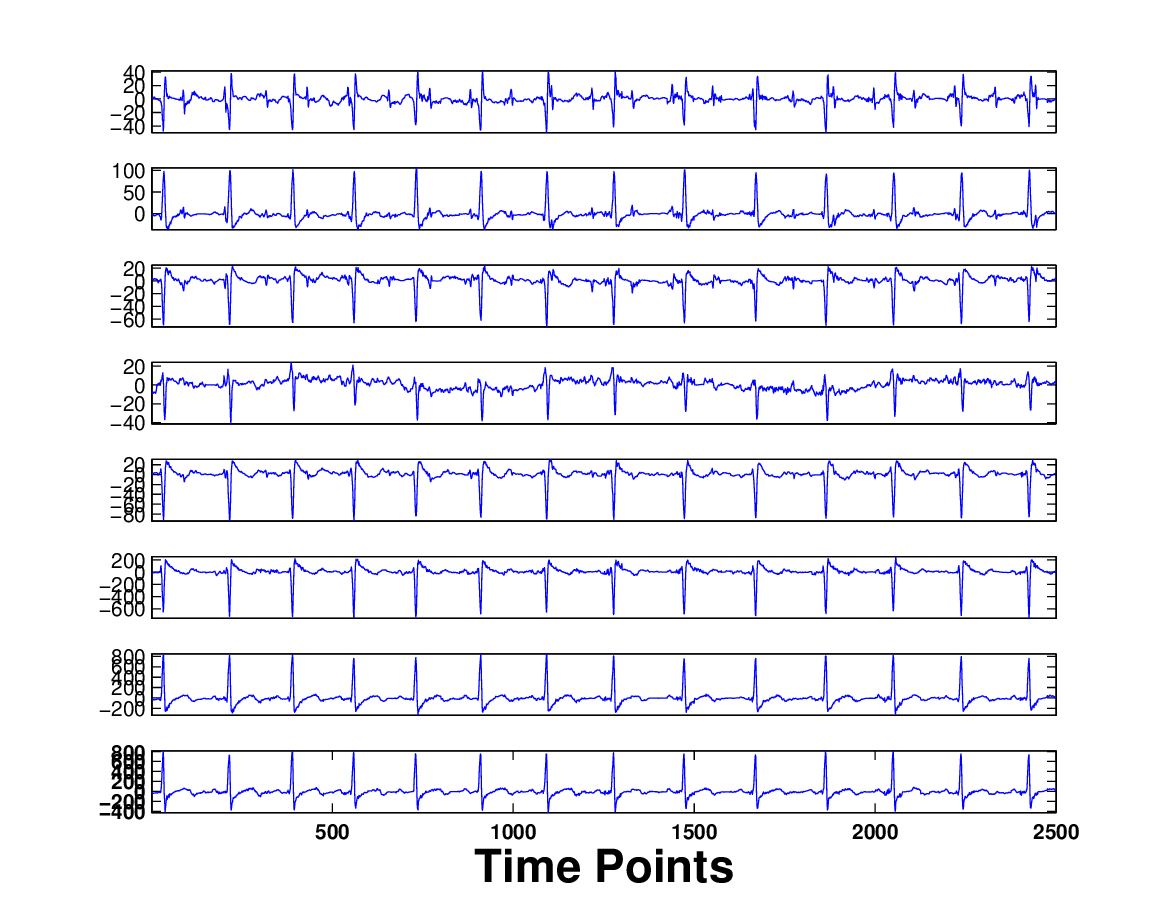,width=4.8cm,height=5cm}}
  \centerline{\footnotesize{(b) Reconstructed Dataset}} \hfill
\end{minipage}
\begin{minipage}[b]{0.48\linewidth}
  \centering
  \centerline{\epsfig{figure=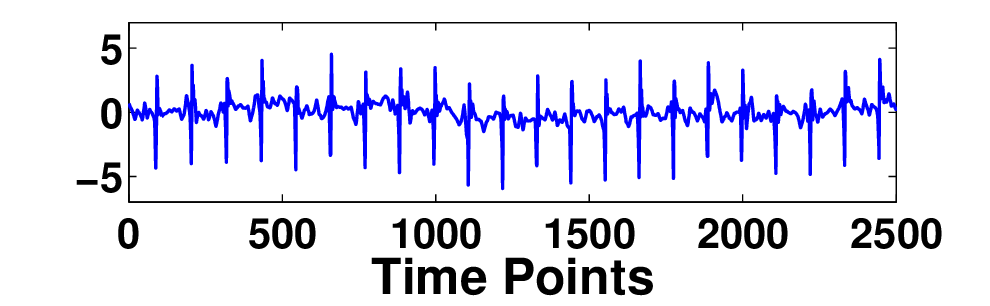,width=4.8cm,height=1.5cm}}
  \centerline{\footnotesize{(c) FECG from (a)}}
\end{minipage}
\hfill
\begin{minipage}[b]{0.48\linewidth}
  \centering
  \centerline{\epsfig{figure=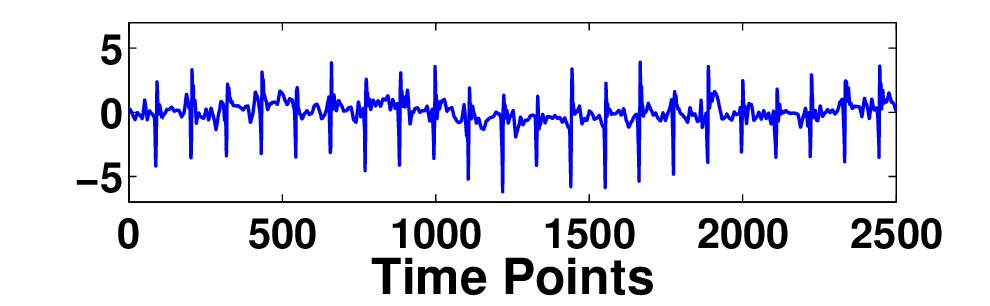,width=4.8cm,height=1.5cm}}
  \centerline{\footnotesize{(d) FECG from (b)}}
\end{minipage}
\caption{(a) The original dataset. (b) The reconstructed dataset by BSBL-BO. (c) The extracted FECG from the original dataset. (d) The extracted FECG from the dataset reconstructed by BSBL-BO.}
\label{fig:ecg_wholedata}
\end{figure}

Then we employed two groups of CS algorithms. One group was the algorithms based on the basic CS model (\ref{equ:SMV model_noiseless}), which do not exploit block structure of signals. They were CoSaMP \cite{Cosamp}, Elastic-Net \cite{zou2005regularization}, Basis Pursuit \cite{van2008probing}, SL0 \cite{SL0}, and EM-GM-AMP \cite{vila2011expectation} (with the `heavy-tailed' mode). They are representative of greedy algorithms, of algorithms minimizing the combination of $\ell_1$ and $\ell_2$ norms, of algorithms minimizing $\ell_1$ norm, of algorithms minimizing $\ell_0$ norm, and of message passing algorithms, respectively. Note that the Basis Pursuit algorithm was the one used in \cite{mamaghanian2011compressed} to reconstruct adult ECG recordings. Their reconstruction results are shown in Figure \ref{fig:exp1_CS} (a)-(e) \footnote{The free parameters of these algorithms were tuned by trial and error. But no values were found to give meaningful results.}.

The second group was the algorithms exploiting structure of signals. They were Block-OMP \cite{BOMP}, Block Basis Pursuit \cite{van2008probing}, CluSS-MCMC \cite{bcs_mcmc}, StructOMP \cite{Huang2009}, and BM-MAP-OMP \cite{faktor2010exploiting}. Block-OMP and Block Basis Pursuit need \emph{a priori} knowledge of the block partition. We used the block partition (\ref{equ:partition}) with $h_1=\cdots=h_g=h$, and $h$ varied from 2 to 50. However, no block sizes yielded meaningful results. Figure \ref{fig:exp1_CS} (f)-(g) display their results when $h=25$.  Figure \ref{fig:exp1_CS} (h) shows the reconstruction result of CluSS-MCMC. StructOMP requires \emph{a priori} knowledge of the sparsity (i.e. the number of nonzero entries in the segment). Since we did not know the sparsity exactly, we set the sparsity from 50 to 250. However, no sparsity value led to a good result. Figure \ref{fig:exp1_CS} (i) shows the result with the sparsity set to 125. Figure \ref{fig:exp1_CS} (j) shows the result of BM-MAP-OMP.

Comparing all the results we can see only the BSBL-BO algorithm, if allowed to exploit intra-block correlation, reconstructed the segment with satisfactory quality.

To further verify the ability of BSBL-BO, we used the same sensing matrix to compress the whole DaISy dataset, and then used BSBL-BO to reconstruct it.

Figure \ref{fig:ecg_wholedata} (a) shows the whole dataset. The most obvious activity is the MECG, which can be seen in all the recordings. The FECG is very weak, which is nearly discernible in the first five recordings. The fourth recording is dominated by a baseline wander probably caused by maternal respiration.

The reconstruction result by BSBL-BO is shown in Figure \ref{fig:ecg_wholedata} (b). All the recordings were reconstructed well. Visually, we do not observe any distortions in the reconstructed dataset.

Admittedly, the reconstructed recordings contained small errors. Since the final goal in our application is to extract clean FECGs from reconstructed FECG recordings using advanced signal processing techniques such as BSS/ICA, we should study whether the reconstruction errors deteriorate the performance of these techniques when extracting FECGs. Here, we examined whether the errors affected the performance of BSS/ICA. We used the eigBSE algorithm, a BSS algorithm proposed in \cite{zhang2006robust}, to extract a clean FECG from the reconstructed recordings. The algorithm exploits quasi-periodic characteristics of FECGs. Thus, if the quasi-periodic structure of FECGs and the ICA mixing structure of the recordings are distorted, the extracted FECG will have poor quality.

Figure \ref{fig:ecg_wholedata} (d) shows the extraction result. We can see the FECG was clearly extracted without losing any QRS complexes or containing residual noise/interference. For comparison, we performed the eigBSE algorithm on the original recordings to extract the FECG. The result is shown in Figure \ref{fig:ecg_wholedata} (c). Obviously, the two extracted FECGs were almost the same. In fact, their Pearson correlation was 0.931.

\subsection{The OSET Dataset}
\label{subsec:data2}

\begin{figure}[tbp]
\begin{minipage}[b]{.48\linewidth}
  \centering
  \centerline{\epsfig{figure=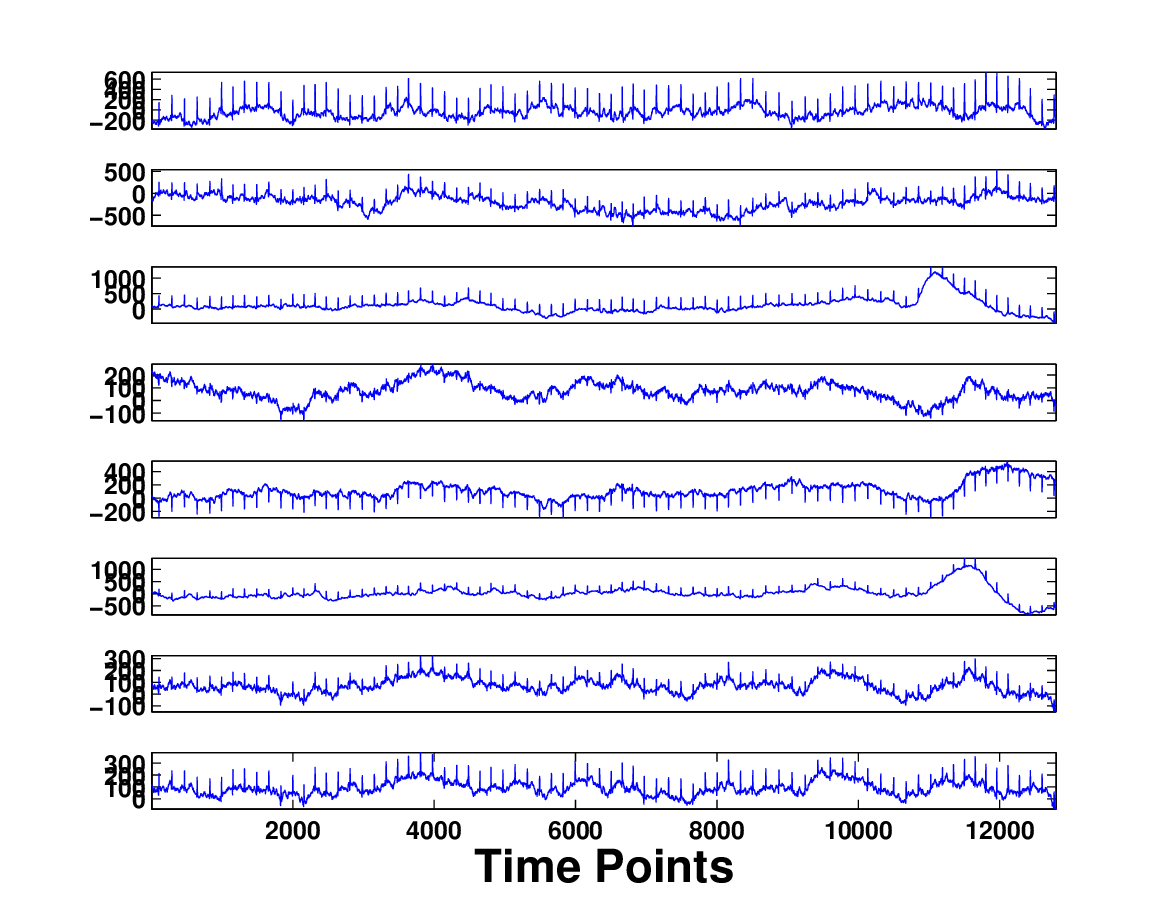,width=4.8cm,height=5cm}}
  \centerline{\footnotesize{(a) Original OSET Dataset}}
\end{minipage}
\hfill
\begin{minipage}[b]{0.48\linewidth}
  \centering
  \centerline{\epsfig{figure=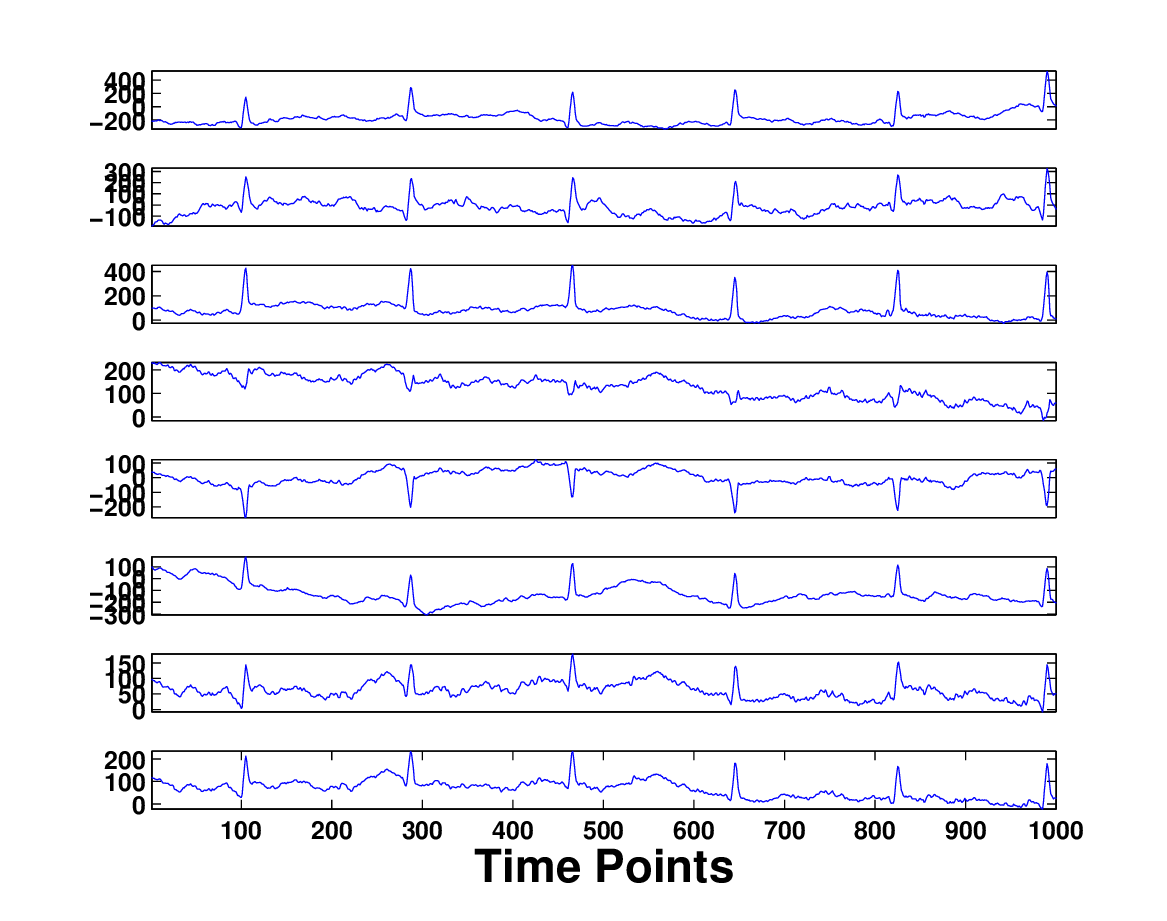,width=4.8cm,height=5cm}}
  \centerline{\footnotesize{(b) First 1000 Time Points}}
\end{minipage}
\caption{The original dataset used in Section \ref{subsec:data2}. (a) The whole dataset, which contains strong baseline wanders. (b) The close-up of the first 1000 time points of the recordings, where only the QRS complexes of the MECG can be observed. The QRS complexes of the FECG are not visible.}
\label{fig:exp2_dataset_original}
\end{figure}

Generally in raw FECG recordings there are many strong baseline wanders, and FECGs are very weak and are buried by noise or MECGs. To test the ability of BSBL-BO in these worse scenarios, we used the dataset `signal01' in the Open-Source Electrophysiological Toolbox (OSET) \cite{OSET}. The dataset consists of eight abdominal recordings sampled at 1000 Hz. We first downsampled the dataset to 250 Hz, since in WBAN-based telemonitoring the sampling frequency rarely exceeds 500 Hz. For illustration, we selected the first 12800 time points of each downsampled recording as the dataset used in our experiment. Figure \ref{fig:exp2_dataset_original} (a) shows the studied dataset, where in every recording the baseline wander is significant. Figure \ref{fig:exp2_dataset_original} (b) shows the first 1000 time points of the recordings, where the QRS complexes of the MECG and various kinds of noise dominate the recordings and the FECG is completely buried by them.

\begin{figure}[tbp]
\begin{minipage}[b]{.48\linewidth}
  \centering
  \centerline{\epsfig{figure=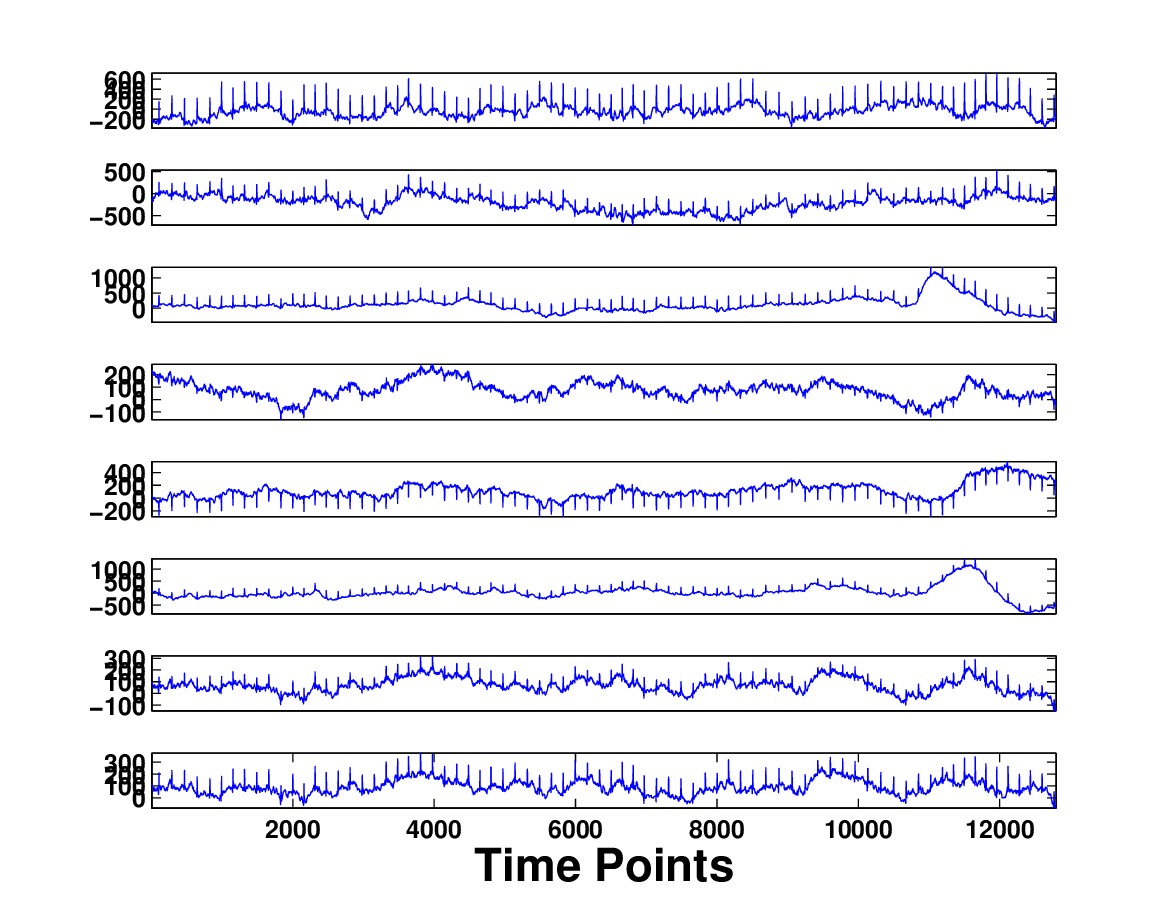,width=4.8cm,height=5cm}}
  \centerline{\footnotesize{(a) Recovered Dataset by BSBL-BO}}
\end{minipage}
\hfill
\begin{minipage}[b]{0.48\linewidth}
  \centering
  \centerline{\epsfig{figure=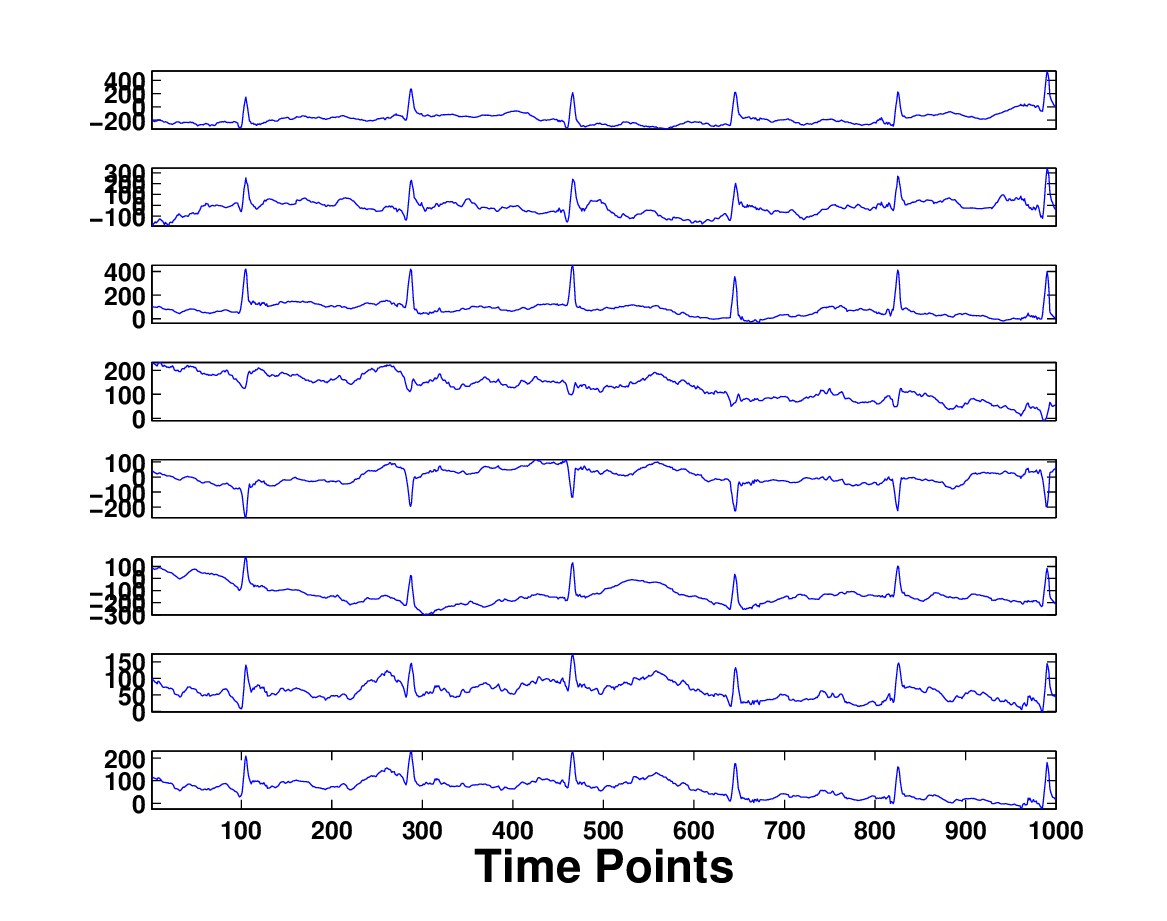,width=4.8cm,height=5cm}}
  \centerline{\footnotesize{(b) First 1000 Time Points}}
\end{minipage}
\caption{The recovered dataset by BSBL-BO. (a) The recovered whole dataset. (b) The first 1000 time points of the recovered dataset. }
\label{fig:exp2_dataset_recovery}
\end{figure}

We used another randomly generated sparse binary sensing matrix of the size $256 \times 512$ with each column consisting of 12 entries of 1s with random locations, while other entries were all zero. The sensing matrix is exactly the one used in \cite{mamaghanian2011compressed}.

For BSBL-BO, we set the block partition $h_1=\cdots=h_{16}=32$. The recovered dataset by BSBL-BO is shown in Figure \ref{fig:exp2_dataset_recovery} (a), and the first 1000 time points of the recovered recordings are shown in Figure \ref{fig:exp2_dataset_recovery} (b). Visually, the recovered dataset was the same as the original dataset, even the baseline wanders were recovered well.

\begin{figure}[tbp]
\begin{minipage}[b]{.48\linewidth}
  \centering
  \centerline{\epsfig{figure=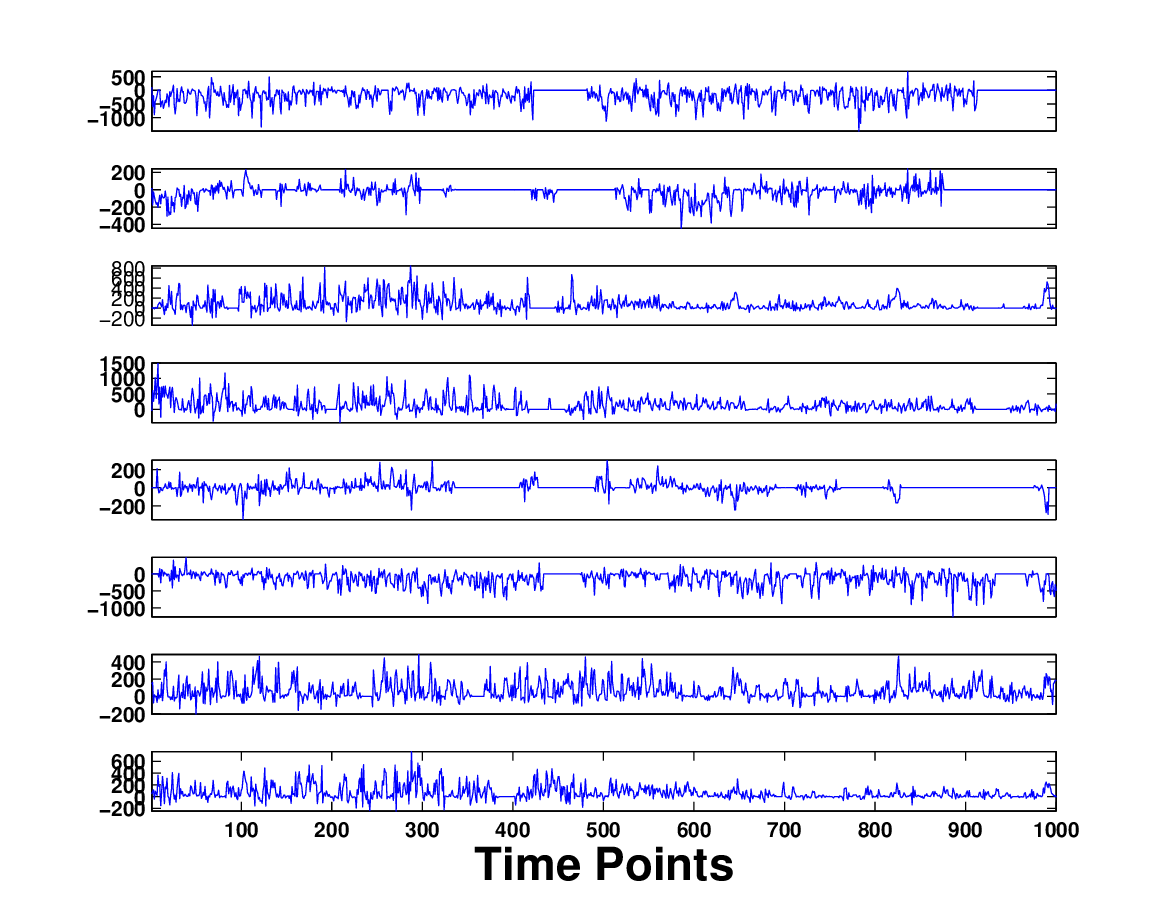,width=4.9cm,height=4.5cm}}
  \centerline{\footnotesize{(a) Recovered by CluSS-MCMC}}
\end{minipage}
\hfill
\begin{minipage}[b]{0.48\linewidth}
  \centering
  \centerline{\epsfig{figure=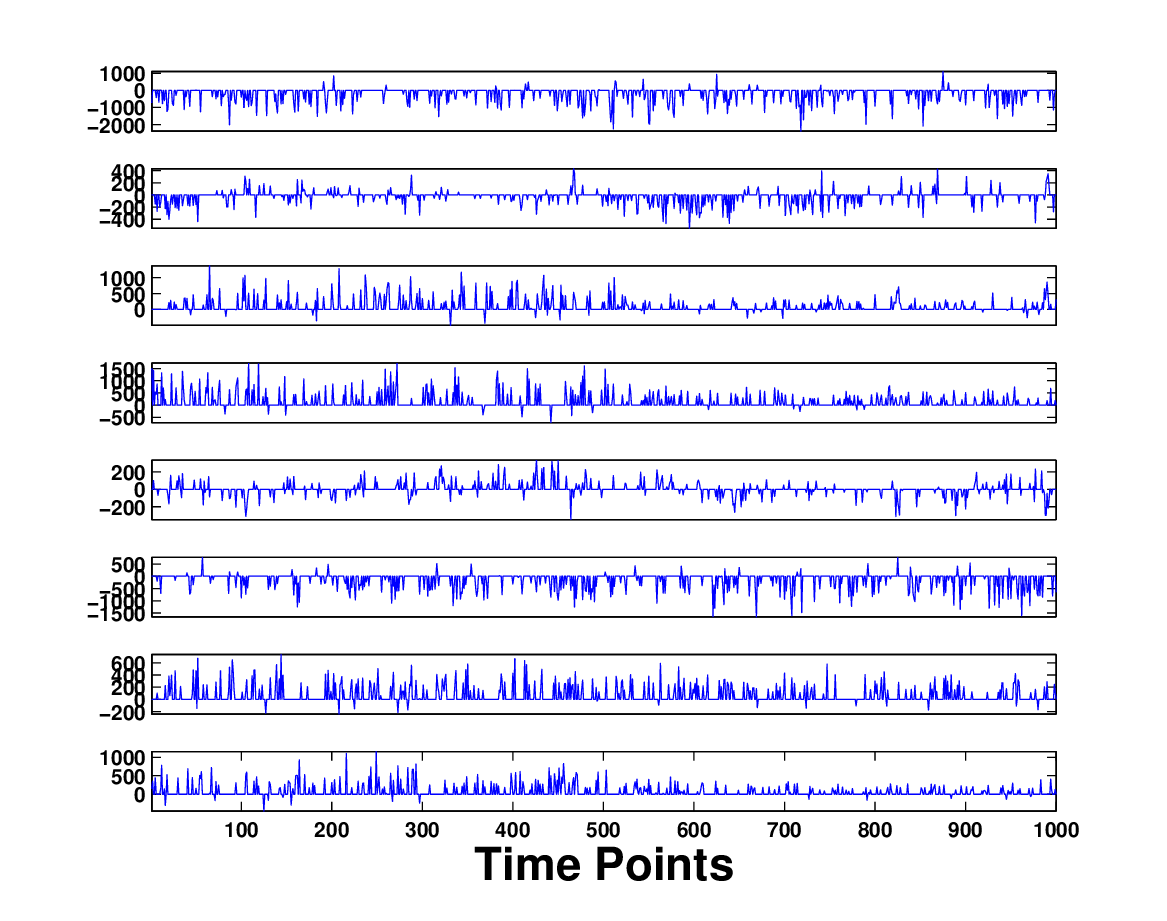,width=4.9cm,height=4.5cm}}
  \centerline{\footnotesize{(b) Recovered by BM-MAP-OMP}}
\end{minipage}
\caption{The whole datasets recovered by (a) CluSS-MCMC and (b) BM-MAP-OMP, respectively.}
\label{fig:exp2_CS_timedomain}
\end{figure}

The previous ten CS algorithms were used to reconstruct the dataset. Again, they all failed. To save space, only the results by CluSS-MCMC and BM-MAP-OMP are presented (Figure \ref{fig:exp2_CS_timedomain}).

Similar to the previous subsection, we used BSS/ICA to extract the FECG and then compared it to the one extracted from the original dataset. Here we used another ICA algorithm, the FastICA algorithm \cite{fastica}.

\begin{figure}[t]
\begin{minipage}[b]{0.48\linewidth}
  \centering
  \centerline{\epsfig{figure=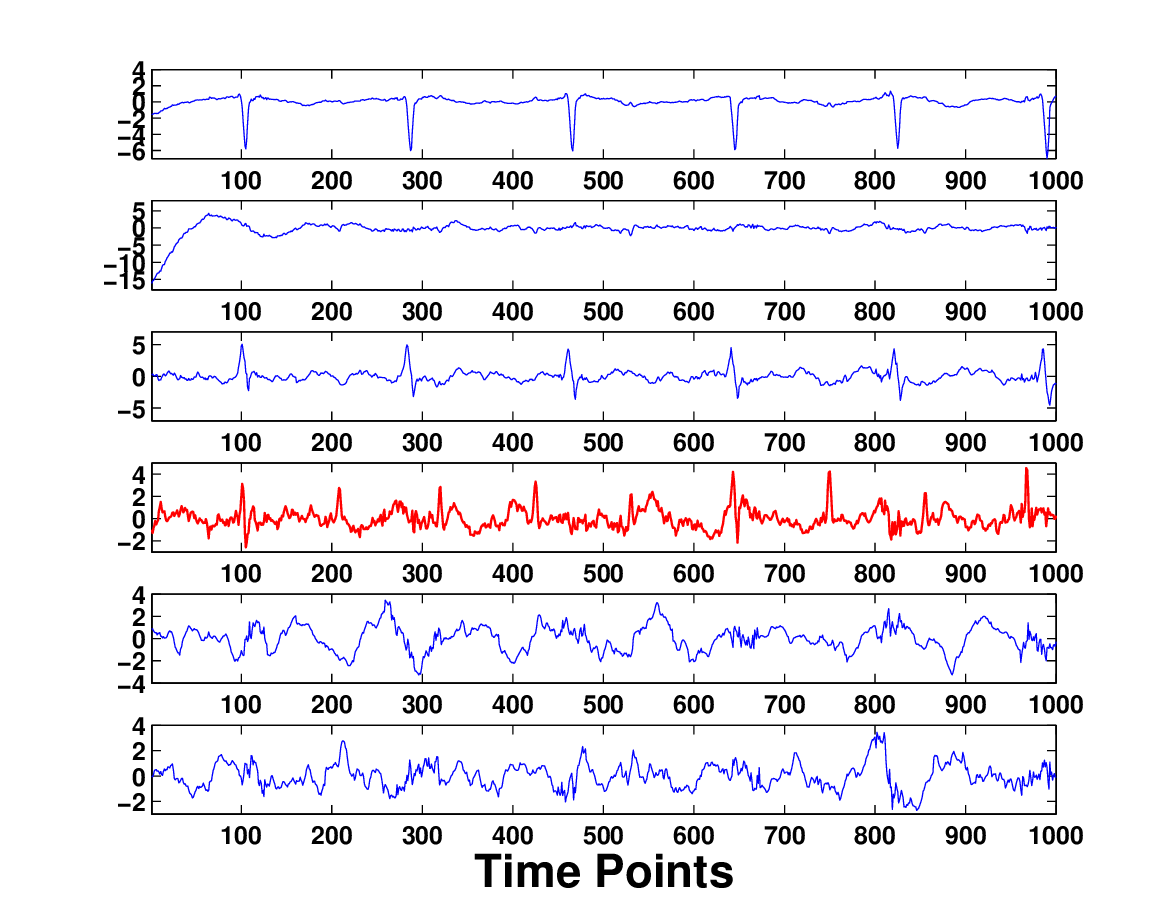,width=4.6cm,height=5.6cm}}
  \centerline{\footnotesize{(a) ICA of the Reconstructed Dataset}}
\end{minipage}
\begin{minipage}[b]{.48\linewidth}
  \centering
  \centerline{\epsfig{figure=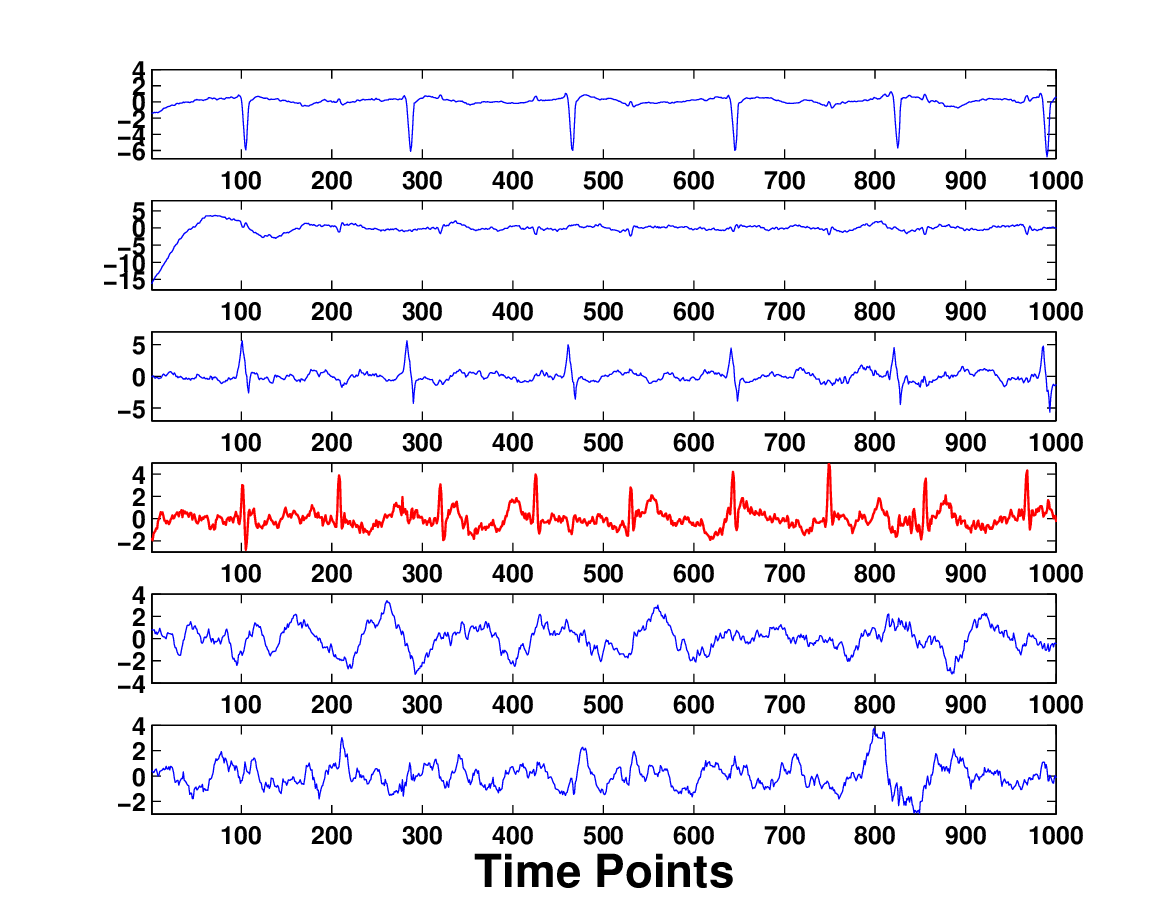,width=4.6cm,height=5.6cm}}
  \centerline{\footnotesize{(b) ICA of the Original Dataset}}
\end{minipage}
\caption{Comparison of the ICA decomposition of the original dataset and the recovered dataset by BSBL-BO. (a) The ICs of the recovered dataset. (b) The ICs of the original dataset. The fourth ICs in (a) and (b) are the extracted FECGs from the reconstructed dataset and the original dataset, respectively.}
\label{fig:exp2_ICA}
\end{figure}

First, the reconstructed dataset was band-passed from 1.75 Hz to 100 Hz (note that in telemonitoring, it is done in the reconstruction stage in remote terminals). Then, FastICA was performed in the `deflation' mode. Six independent components (ICs) with significant non-Gaussianity were extracted, as shown in Figure \ref{fig:exp2_ICA} (a), where the fourth IC is the FECG.

Then FastICA was performed on the original dataset. The ICs are shown in Figure \ref{fig:exp2_ICA} (b). Comparing Figure \ref{fig:exp2_ICA} (a) with Figure \ref{fig:exp2_ICA} (b) we can see the distortion was very small, which obviously did not affect clinical diagnosis.

\begin{figure}[tbp]
\begin{minipage}[b]{0.48\linewidth}
  \centering
  \centerline{\epsfig{figure=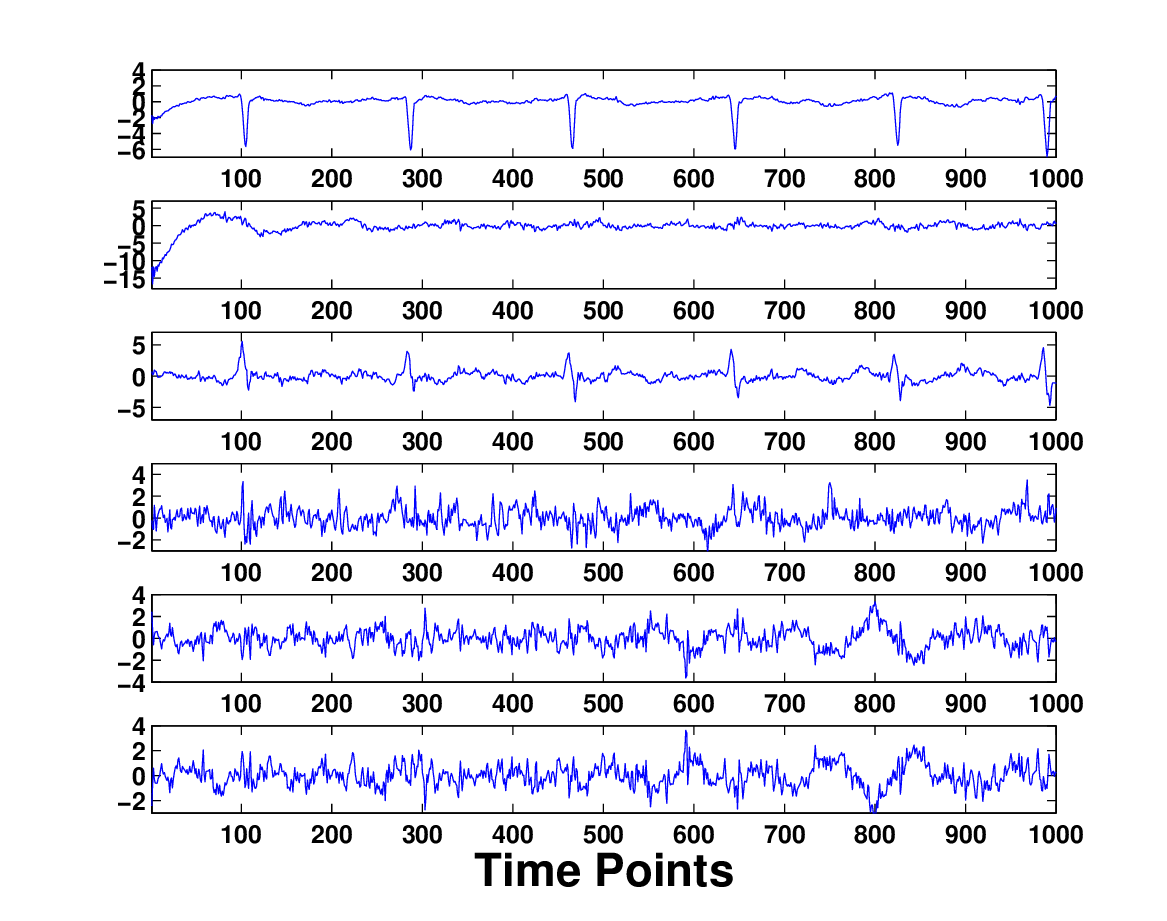,width=4.6cm,height=6cm}}
  \centerline{\footnotesize{(a)}}
\end{minipage}
\begin{minipage}[b]{.48\linewidth}
  \centering
  \centerline{\epsfig{figure=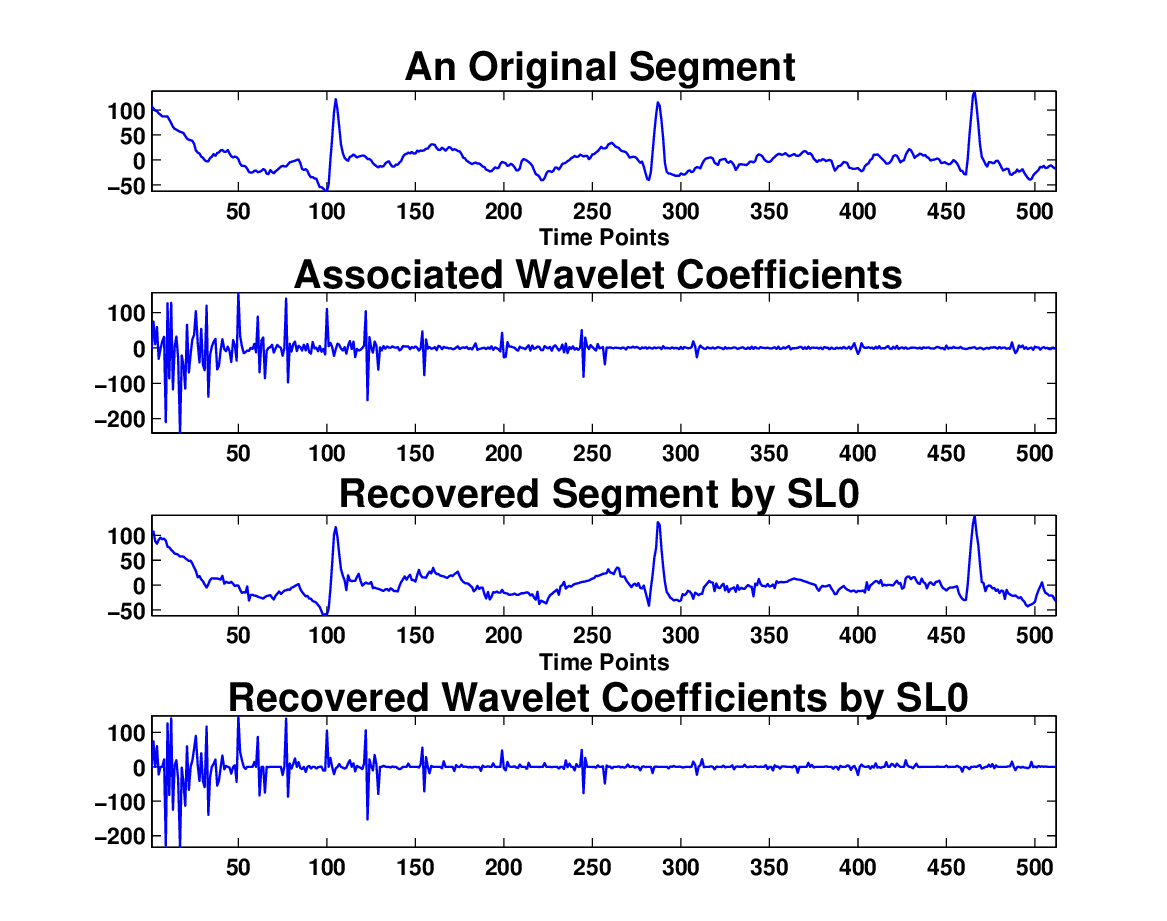,width=4.6cm,height=6cm}}
  \centerline{\footnotesize{(b)}}
\end{minipage}
\caption{Reconstruction result by SL0 with the aid of the wavelet transform. (a) The ICs from the recovered dataset by SL0. (b) From top to bottom are a segment of the original dataset, the associated wavelet coefficients, the recovered segment by SL0, and the recovered wavelet coefficients by SL0. }
\label{fig:exp2_SL0}
\end{figure}

\subsection{Reconstruction in the Wavelet Domain}

To reconstruct non-sparse signals, a conventional approach in the CS field is to adopt the model (\ref{equ:CS_model}), namely, first reconstructing $\boldsymbol{\theta}$ using the received data $\mathbf{y}$ and the known matrix $\mathbf{\Omega}$, and then calculating $\mathbf{x}$ by $\mathbf{x}=\mathbf{\Psi} \boldsymbol{\theta}$. To test whether this approach is helpful for existing CS algorithms to reconstruct raw FECG recordings, in the following we repeated the experiment in Section \ref{subsec:data2} using the previous ten CS algorithms and this approach.

Since it is suggested \cite{eduardo2010implementation} that Daubechies-4 wavelet can yield very sparse representation of ECG, we set $\mathbf{\Psi}$ to be the orthonormal basis of Daubechies-4 wavelet. The sensing matrix was the one used in Section \ref{subsec:data2}.

Unfortunately, all these CS algorithms failed again. The FECG was not extracted from the dataset reconstructed by any of these CS algorithms. Figure \ref{fig:exp2_SL0} (a) shows the ICs extracted from the dataset reconstructed by SL0 based on the wavelet basis. Obviously, the FECG was not extracted.

Therefore, using the wavelet transform is still not helpful for these CS algorithms. The reason is that to ensure the FECG can be extracted by ICA with high fidelity, the ICA mixing structure should be maintained well in the reconstructed dataset. This requires that wavelet coefficients with small amplitudes in addition to those with large amplitudes are all recovered well. However, for a raw FECG recording the number of wavelet coefficients with small amplitudes is very large. To recover these coefficients is difficult for the CS algorithms.

As an example, the top two panels in Figure \ref{fig:exp2_SL0} (b) show a segment of a raw recording and its wavelet coefficients, respectively. The bottom two panels in Figure \ref{fig:exp2_SL0} (b) show the recovered segment and the recovered wavelet coefficients by SL0, respectively. We can see the coefficients with large amplitudes were recovered well. However, it failed to recover the coefficients with small amplitudes, which resulted in the failure of ICA to extract the FECG.

\section{Performance Issues of BSBL-BO}
\label{sec:issues}

This section explores how the performance of BSBL-BO is affected by various experimental factors.

\subsection{Effects of Signal-to-Interference-and-Noise Ratio}

We have tested BSBL-BO's performance using two typical datasets. The two datasets contain MECGs and noise with certain strength. It is natural to ask whether BSBL-BO can be used for other datasets containing MECGs and noise with different strength. This question is very important, since different fetus positions, different pregnancy weeks, and random muscle movements can result in dramatic changes in   correlation structure of raw recordings, while BSBL-BO exploits the correlation structure to improve performance.

Therefore, we carried out Monte Carlo simulations with different strength of FECGs, MECGs, and other noise, as in \cite{sameni2010deflation}. The raw multichannel recordings were modeled as the summation of a multichannel FECG, a multichannel MECG, and  multichannel noise. The multichannel MECG was generated by a three-dimentional dipole which projects cardiac potentials to eight sensors. The multichannel FECG was generated in the same way with half period of the MECG. The angle between the two dipoles generating the FECG potential and the MECG potential was $41^\circ$. The noise was a combination of randomly selected real-world baseline wanders, muscle artifacts, and electrode movement artifacts from the Noise Stress Test Database (NSTDB) \cite{NoiseDatabase}. Details on the simulation design can be found in \cite[Sec. V.A]{sameni2010deflation}. The generated raw recordings were downsampled to 250 Hz. Each recording finally contained 7680 time points.

As in \cite{sameni2010deflation}, the ratio of the power of the multichannel FECG to the power of the multichannel MECG was defined as the Signal-to-Interference Ratio (SIR). The ratio of the power of the multichannel FECG to the power of the multichannel noise was defined as the Signal-to-Noise Ratio (SNR). And the ratio of the power of the FECG to the combined power of the MECG and the noise  was defined as the Signal-to-Interference-and-Noise Ratio (SINR). In the simulation, the strength of the FECG, the MECG and the noise were adjusted such that SNR = SIR + 10dB, and SINR was swept in the range of -35dB to -15dB. Note that in the experiment the SINR range was intentionally made more challenging, since for most raw recordings the SINR varies only from -5dB to -25dB \cite{shepoval2006investigation}. For each value of the SINR, the simulation was repeated 20 times, each time with different signals and noise.

\begin{figure}[tbp]
\centering
\includegraphics[width=6cm,height=3.5cm]{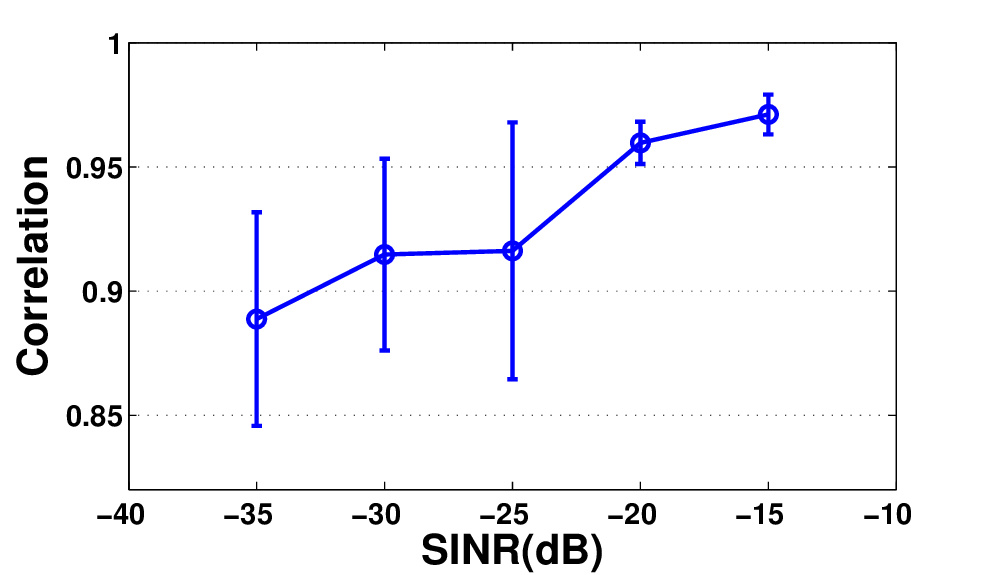}
\caption{The Pearson correlation (averaged over 20 trials) between the extracted FECG from the original dataset and the one from the recovered dataset at different SINRs. The error bar gives the standard variance.}
\label{fig:exp4_SINR}
\end{figure}

\begin{figure}[tbp]
\begin{minipage}[b]{.48\linewidth}
  \centering
  \centerline{\epsfig{figure=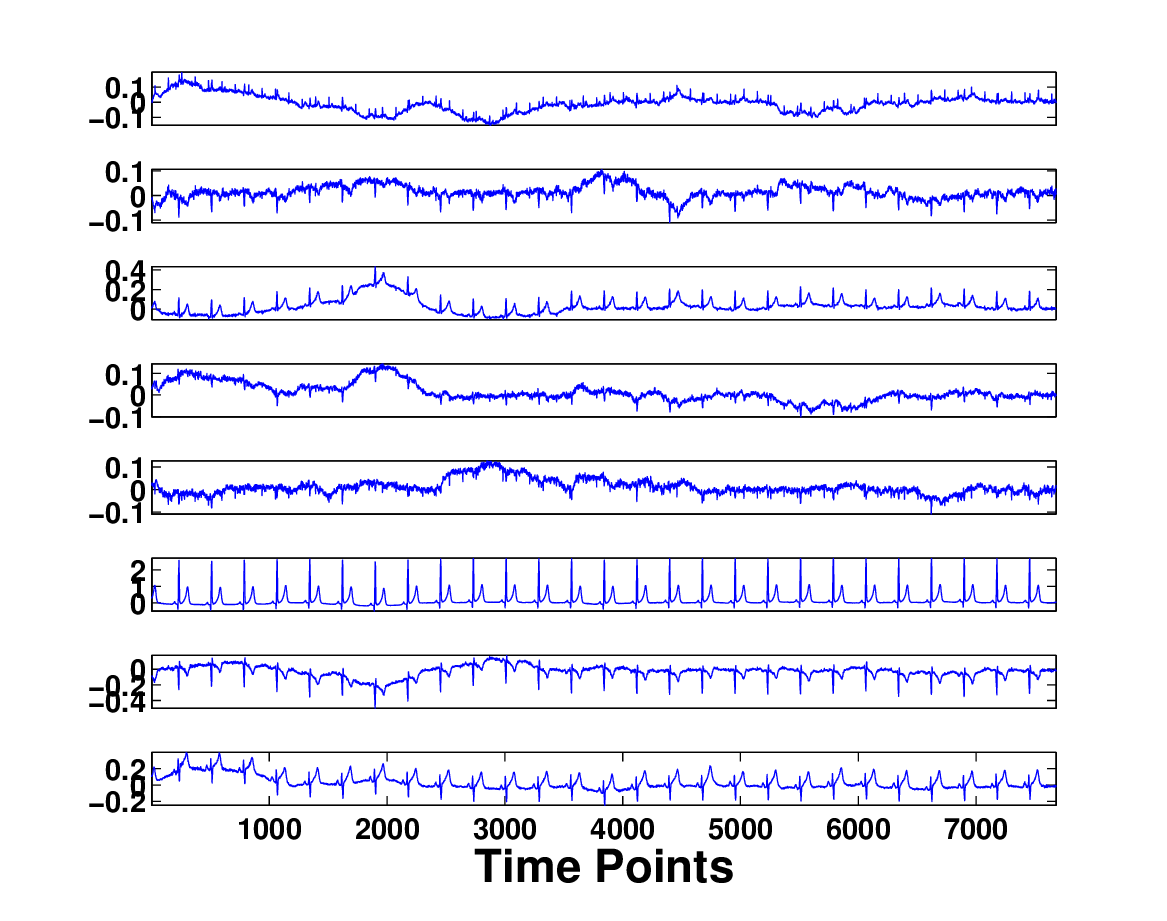,width=4.9cm,height=4.2cm}}
  \centerline{\footnotesize{(a)}}
\end{minipage}
\hfill
\begin{minipage}[b]{0.48\linewidth}
  \centering
  \centerline{\epsfig{figure=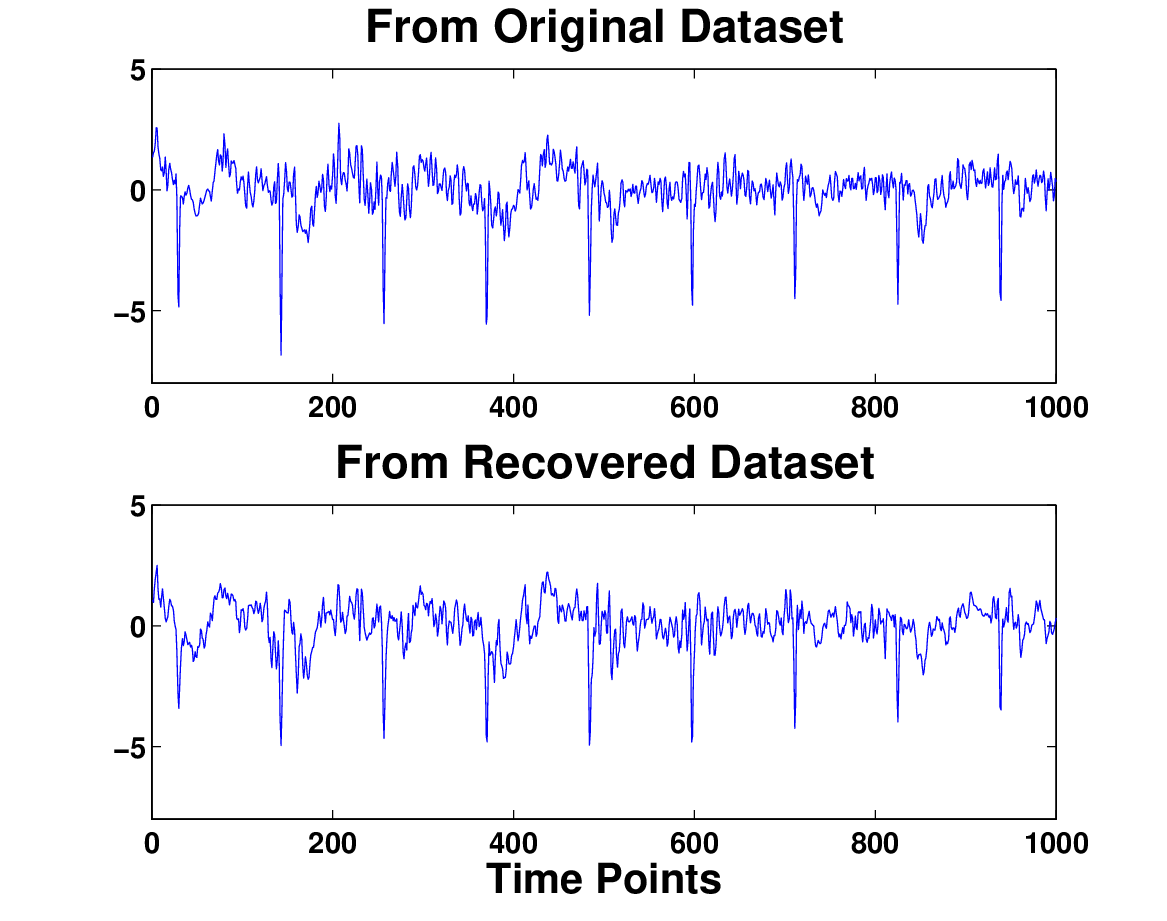,width=4.9cm,height=4.2cm}}
  \centerline{\footnotesize{(b)}}
\end{minipage}
\caption{A synthesized dataset and the extraction result at SINR=-35dB. (a) The synthesized dataset. (b) The comparison between the extracted FECG from the synthesized dataset and the one from the corresponding recovered dataset (only their first 1000 time points are shown).}
\label{fig:exp4}
\end{figure}

The sensing matrix and the block partition of BSBL-BO were the same as in Section \ref{subsec:data2}. The result presented in Figure \ref{fig:exp4_SINR} clearly shows the high recovery quality of BSBL-BO even in the worst scenarios. Figure \ref{fig:exp4} shows a generated dataset when SINR=-35dB, and the extracted FECGs from the generated dataset and from the recovered dataset. We can see the noise was very strong, but the extracted FECG from the recovered dataset still maintained high fidelity.

\begin{figure}[tp]
\centering
\includegraphics[width=6cm,height=4.2cm]{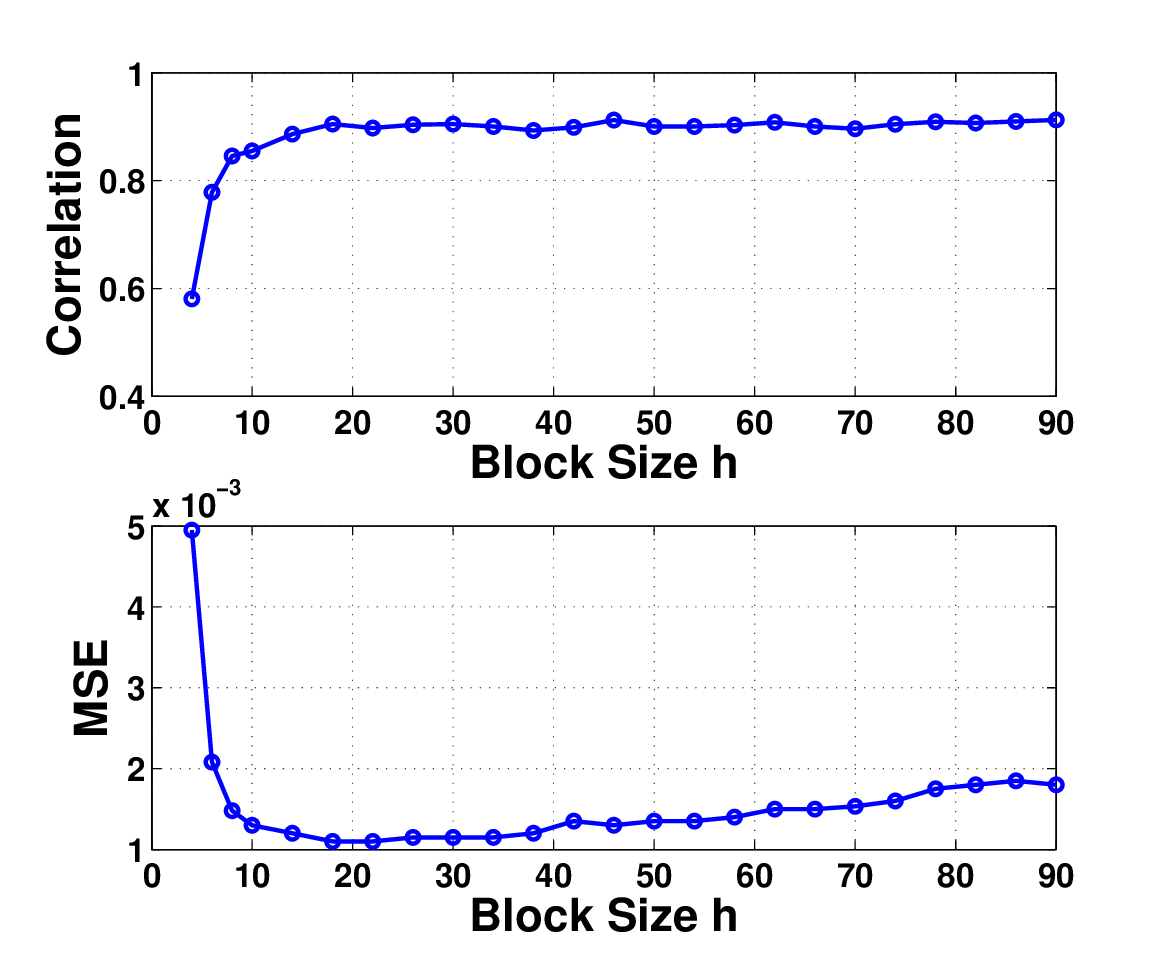}
\caption{Effects of the block size $h$ on the reconstruction quality, measured by the correlation between the extracted FECG from the reconstructed dataset and the extracted one from the original dataset (upper panel), and by the MSE of the reconstructed dataset (bottom panel). }
\label{fig:blocksize}
\end{figure}

\subsection{Effects of the Block Partition}

In all the previous experiments we used certain block partitions. Another question is, ``Is the performance of BSBL-BO sensitive to the block partition?" To examine this, we used the  dataset in Section \ref{subsec:data2}. The block partition was designed as follows: the location of the first entry of each block was $1, 1+h, 1+2h, \cdots$, respectively, where the block size $h$ ranged from 4 to 90. The sensing matrix was a sparse binary matrix of the size $128 \times 256$. Its each column contained 12 nonzero entries of 1s with random locations. The experiment was repeated 20 times. In each time the sensing matrix was different.

The averaged results are shown in Fig.\ref{fig:blocksize}, from which we can see that the extraction quality was almost the same over a broad range of $h$.

\subsection{Effect of Compression Ratio}
\label{subsec:CR}
Next, we investigated the effect of compression ratio (CR) on the quality of extracted FECGs from reconstructed recordings. The compression ratio is defined as
\begin{eqnarray}
CR = \frac{N-M}{N} \times 100
\end{eqnarray}
where $N$ is the length of the original signal and $M$ is the length of the compressed signal. The used sparse binary sensing matrix was of the size $M \times N$, where $N$ was fixed to 512 and $M$ varied such that CR ranged from 20 to 65. Regardless of the size, its each column contained 12 entries of 1s. For each value of $M$, we repeated the experiment 20 times, and in each time the sensing matrix was randomly generated. The dataset and the block partition for BSBL-BO were the same as in Section \ref{subsec:data2}.

\begin{figure}[tbp]
\begin{minipage}[b]{.48\linewidth}
  \centering
  \centerline{\epsfig{figure=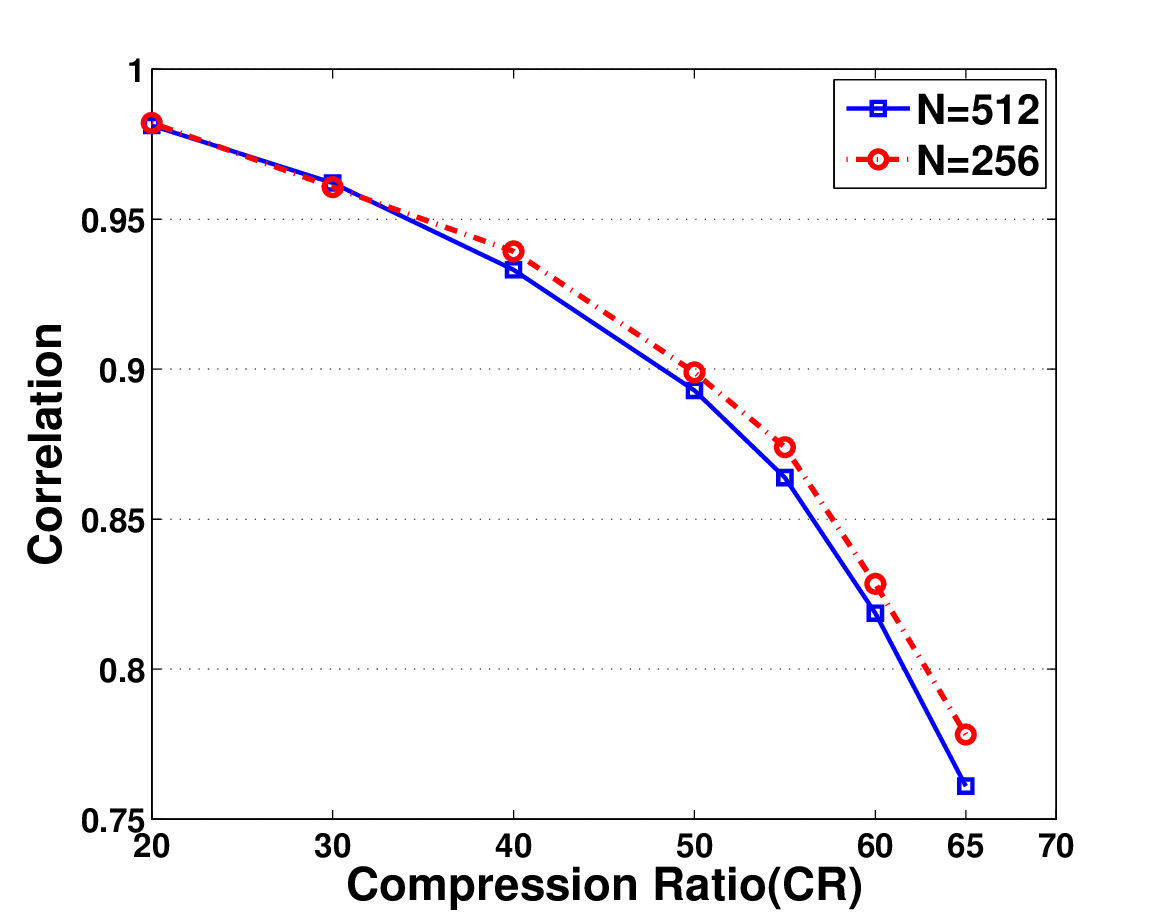,width=4.6cm,height=3.5cm}}
  \centerline{\footnotesize{(a)}}
\end{minipage}
\hfill
\begin{minipage}[b]{0.48\linewidth}
  \centering
  \centerline{\epsfig{figure=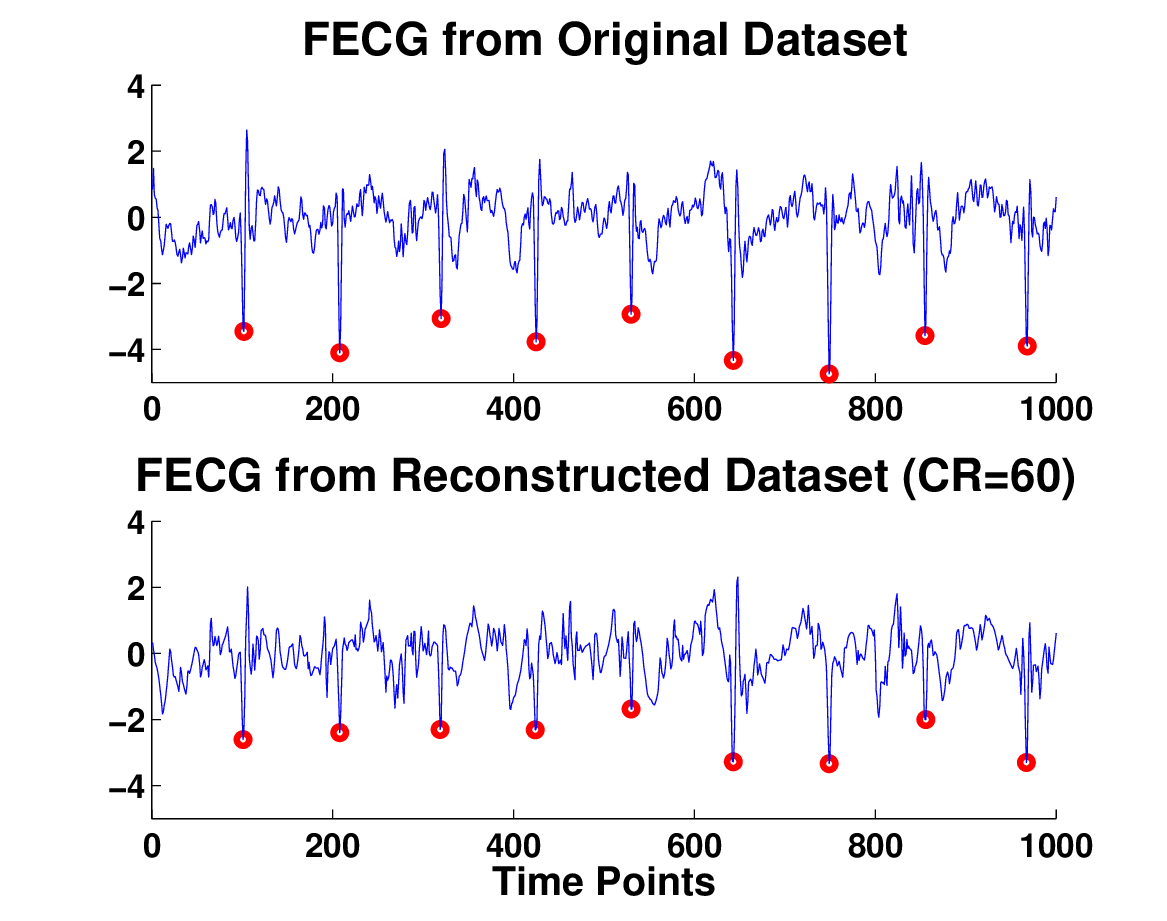,width=4.6cm,height=3.5cm}}
  \centerline{\footnotesize{(b)}}
\end{minipage}
\caption{Experiment results in Section \ref{subsec:CR}. (a) The effect of the compression ratio on the quality of extracted FECGs from reconstructed datasets (measured by correlation) when $N=512$ and $N=256$. (b) The extracted FECG from the original dataset and from the reconstructed dataset when CR=60 and $N=512$ (only first 1000 time points are shown). Red circles indicate the detected peaks of R-waves in both FECGs.}
\label{fig:cr}
\end{figure}

The averaged results for each CR are shown in Figure \ref{fig:cr} (a). We found that when $\mathrm{CR} \leq 60$, the quality of extracted FECGs was satisfactory and could be used for clinical diagnosis. For example, Figure \ref{fig:cr} (b) shows the extracted FECG from a reconstructed dataset when CR=60. Compared to the FECG extracted from the original dataset, the FECG from the reconstructed dataset did not have significant distortion. Especially, when using the `\emph{PeakDetection}' program in the OSET toolbox to detect peaks of the R-waves in both extracted FECGs, the results were the same, as shown in Figure \ref{fig:cr} (b).

We repeated the experiment using a smaller sparse binary matrix with $N=256$. Each column also contained 12 entries of 1s. The block size in the block partition for BSBL-BO did not change. The result (Figure \ref{fig:cr} (a)) shows that the quality of extracted FECGs was slightly better than the case of $N=512$.

\begin{figure}[tbp]
\begin{minipage}[b]{.48\linewidth}
  \centering
  \centerline{\epsfig{figure=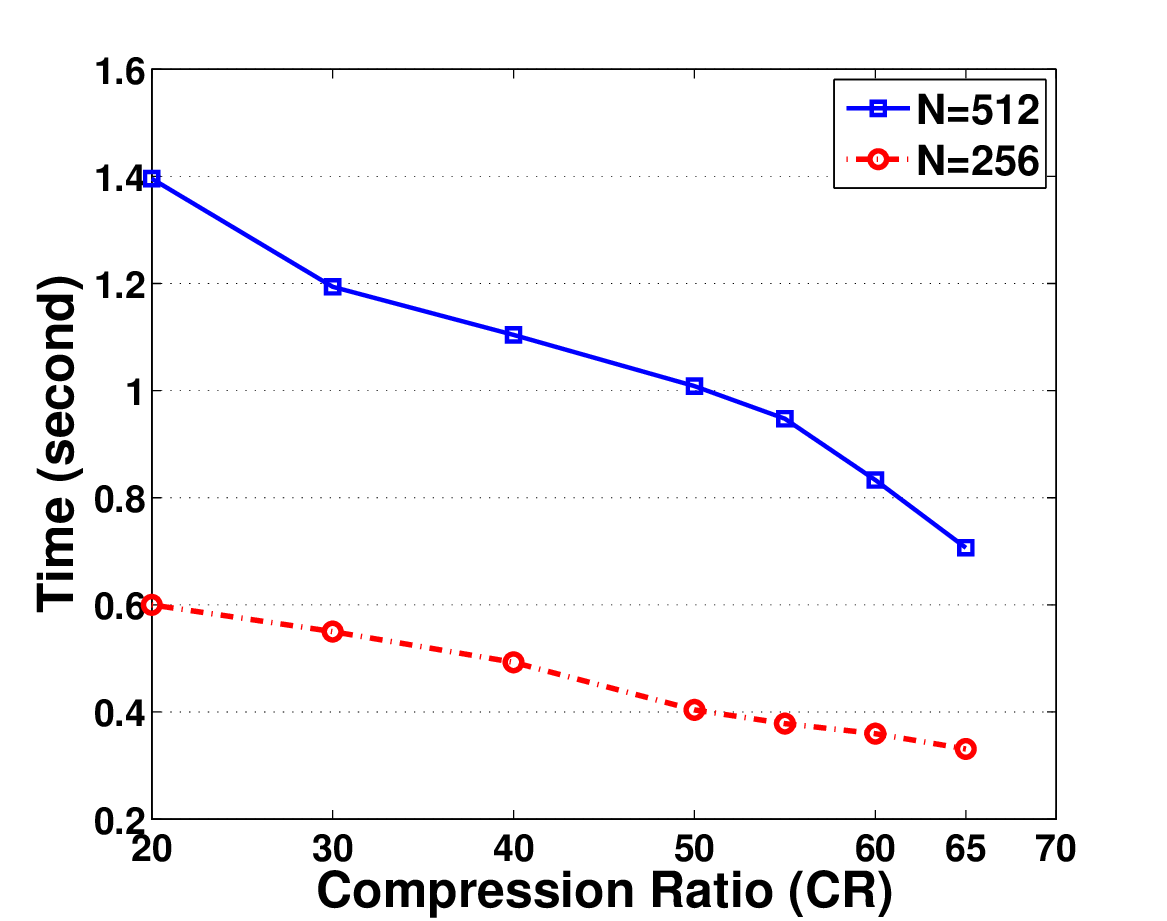,width=4.6cm,height=3.5cm}}
  \centerline{\footnotesize{(a)}}
\end{minipage}
\hfill
\begin{minipage}[b]{0.48\linewidth}
  \centering
  \centerline{\epsfig{figure=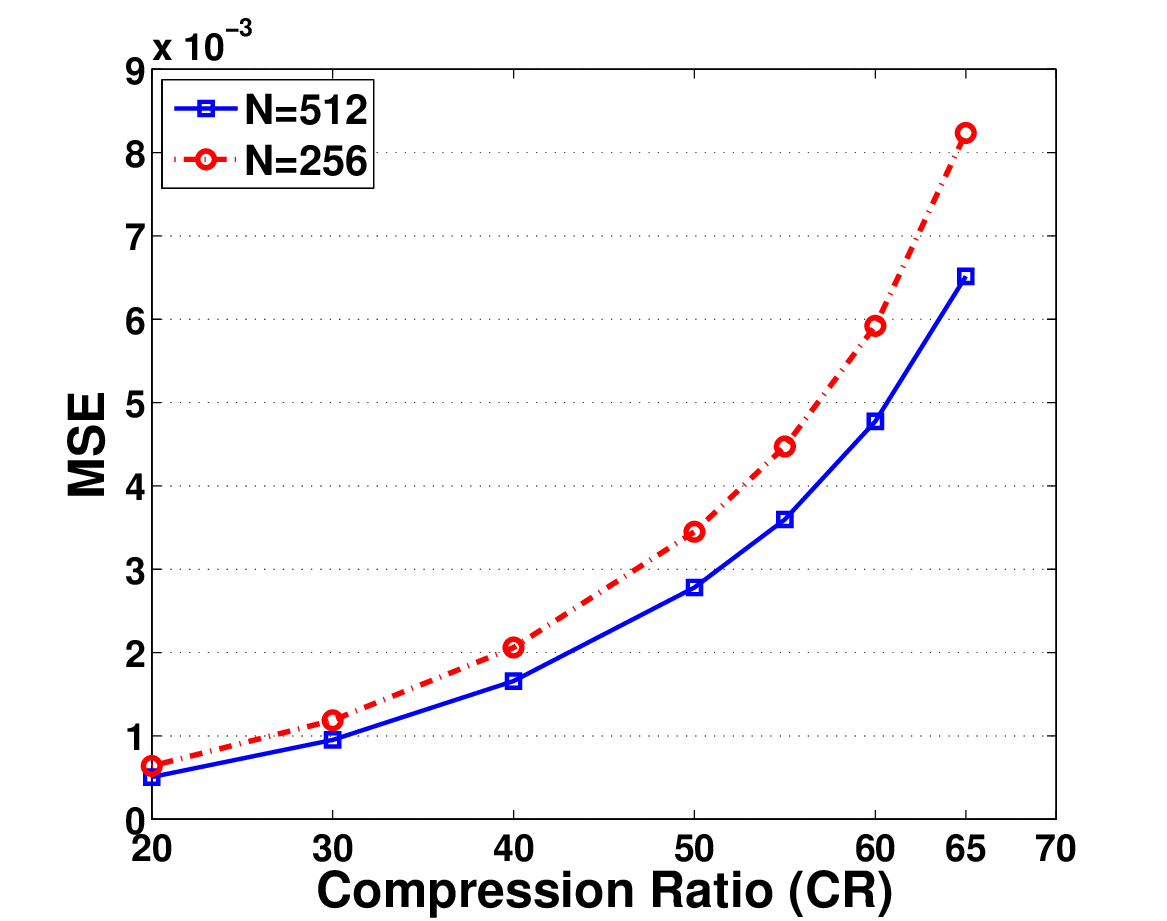,width=4.6cm,height=3.5cm}}
  \centerline{\footnotesize{(b)}}
\end{minipage}
\caption{(a) Comparison of averaged time in reconstructing a segment of 512 time points from the dataset shown in Figure \ref{fig:exp2_dataset_original} when using two sensing matrices. Note that it took 2.048 seconds to collect a segment of 512 time points. But the algorithm took less than 1.4 seconds to recover the segment in a laptop with 2.8G CPU and 6G RAM if using the big sensing matrix, or took less than 0.6 seconds if using the small sensing matrix. (b) Comparison of MSEs in reconstructing the dataset when using the two sensing matrices.}
\label{fig:cr2}
\end{figure}

Note that a significant advantage of using a smaller sensing matrix is that the reconstruction is accelerated. Figure \ref{fig:cr2} (a) compares the averaged time in reconstructing a segment of 512 time points when using the two sensing matrices at different CRs. Clearly, using a small sensing matrix speeded up the reconstruction \footnote{The maximum iteration of BSBL-BO was set to 25.}, making it possible to build a near real-time telemonitoring system.

It is worth noting that when fixing CR, using a smaller sensing matrix generally results in higher MSE of reconstructed recordings, as shown in Figure \ref{fig:cr2} (b). But this does not mean the quality of extracted FECGs is poorer accordingly, as shown in Figure \ref{fig:cr} (a).

\begin{figure}[tp]
\begin{minipage}[b]{.48\linewidth}
  \centering
  \centerline{\epsfig{figure=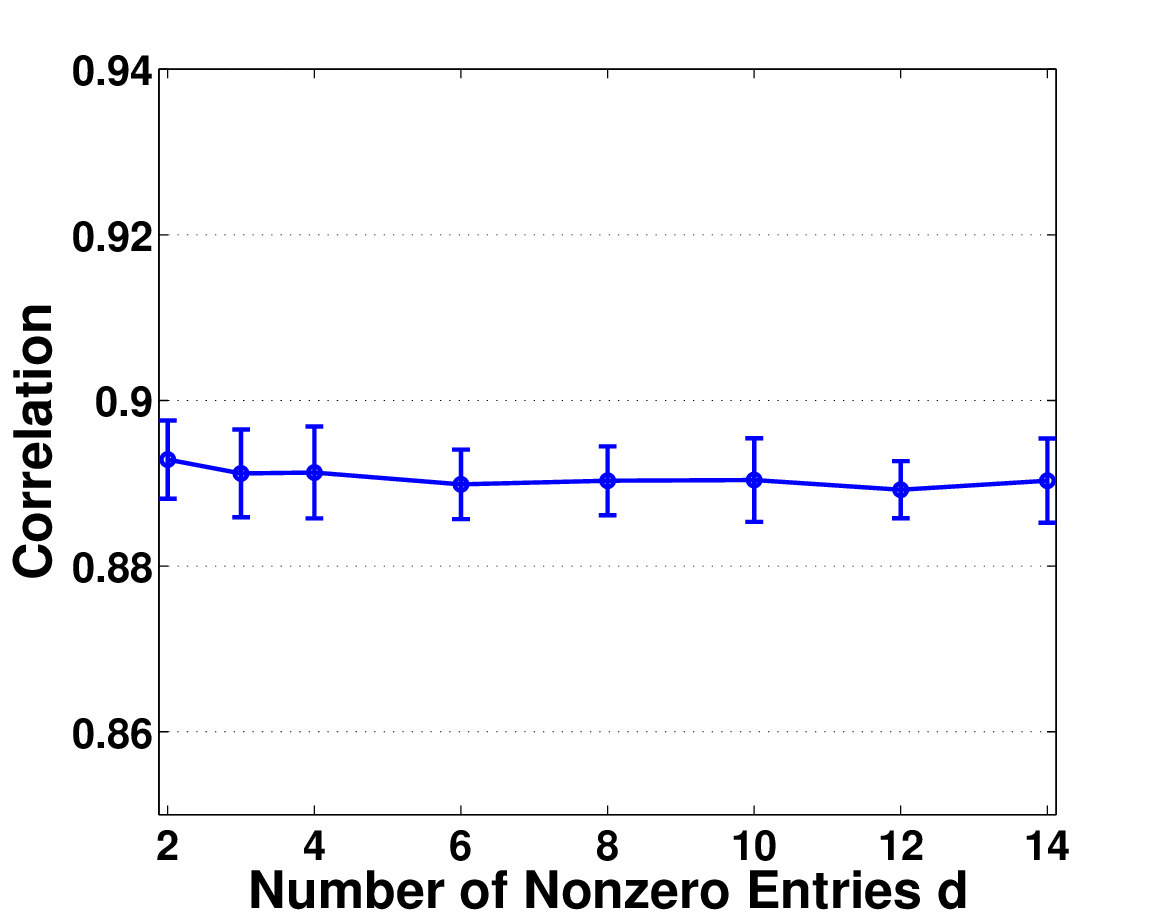,width=4.5cm,height=3.2cm}}
  \centerline{\footnotesize{(a)}}
\end{minipage}
\hfill
\begin{minipage}[b]{0.48\linewidth}
  \centering
  \centerline{\epsfig{figure=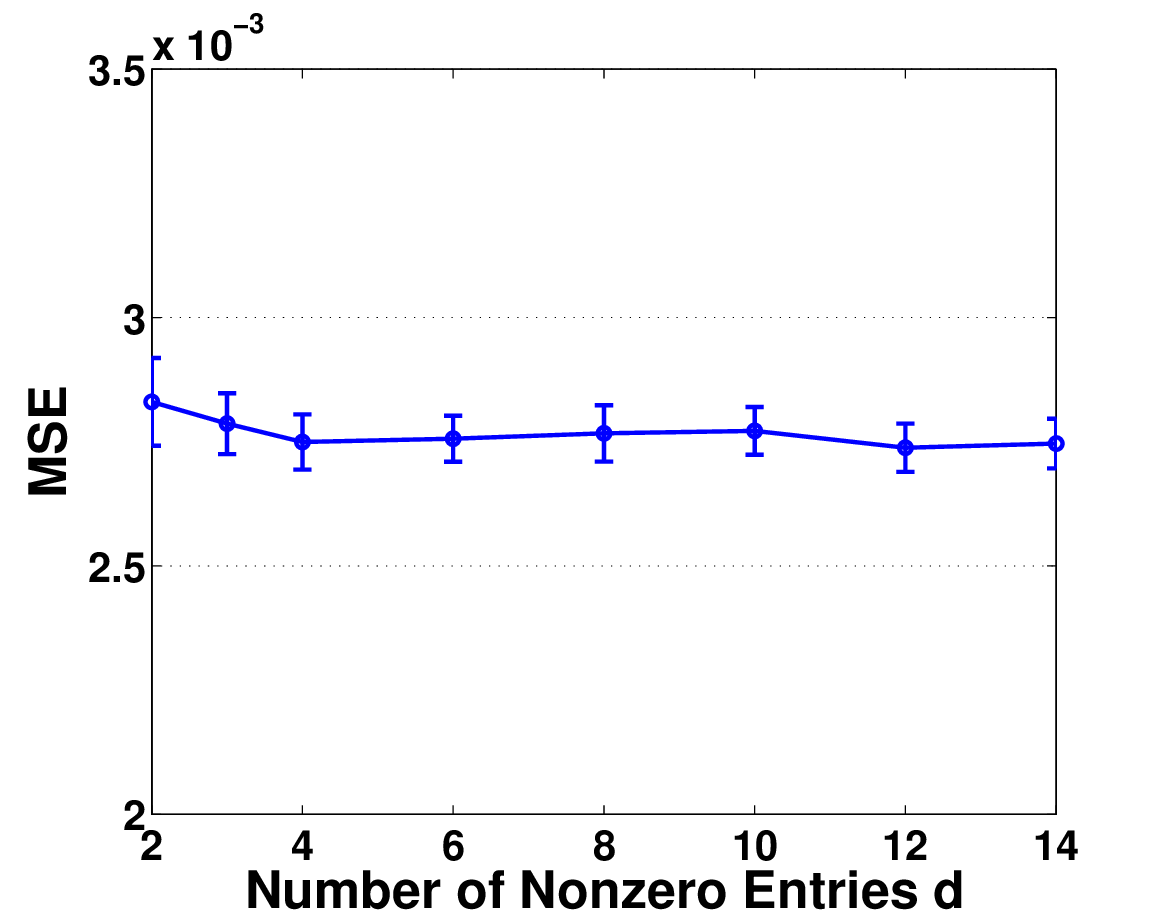,width=4.5cm,height=3.2cm}}
  \centerline{\footnotesize{(b)}}
\end{minipage}
\caption{Relation between recovery quality and the number of entries of 1s in each column of the sensing matrix. In (a) the vertical coordinate gives the correlation between the extracted FECG from the reconstructed dataset and the FECG from the original dataset at different values of $d$. In (b) the vertical coordinate gives the MSEs of the reconstructed datasets. The error bar gives the standard variance.}
\label{fig:exp7_d}
\end{figure}

\subsection{Study on the Number of Nonzero Entries in Each Column of the Sensing Matrix}
\label{subsec:d}

In most experiments we used a $256\times 512$ sensing matrix, and each column contained 12 entries of 1s with random locations. This number of nonzero entries in each column was chosen by Mamaghanian et al in \cite{mamaghanian2011compressed}. To study how the number of nonzero entries in each column affects the performance of BSBL-BO, we carried out a similar experiment as in \cite{mamaghanian2011compressed}.

The sparse binary sensing matrix was of the size $256\times 512$. Each column contained $d$ entries of 1s, where $d$ varied from 2 to 14. The experiment was repeated 20 times for each value of $d$. In each time the locations of the nonzero entries were randomly chosen (but the generated sensing matrix was always full row-rank).  The  dataset and the block partition were the same as in Section \ref{subsec:data2}.

Figure \ref{fig:exp7_d} (a) shows the Pearson correlation between the extracted FECG from the reconstructed dataset and the extracted FECG from the original dataset at different values of $d$. Figure \ref{fig:exp7_d} (b) shows the quality of reconstructed datasets measured by the MSE. Both figures show that the results were not affected by $d$. This is different from the results in \cite{mamaghanian2011compressed}, where a basic $\ell_1$ CS algorithm was used and its performance was very sensitive to $d$.

The robustness to $d$ is another advantage of BSBL-BO, which is important to energy saving, as discussed in Section \ref{subsec:energy}.

\section{Discussions}
\label{sec:discussion}

\subsection{The Block Partition in the BSBL Framework}

The problem of reconstructing a raw FECG recording can be cast as a block sparse model with unknown block partition and unknown signal noise in a noiseless environment. To exploit the unknown block structure, our algorithm is based on a very simple and even counter-intuitive strategy. That is, \emph{the user-defined block partition can be rather arbitrary, which is not required to be the same as the true block structure of the FECG recording}. This strategy is completely different from the strategies used by many CS algorithms to deal with unknown block structure, which try to find the true block structure as accurately as possible \cite{bcs_mcmc,faktor2010exploiting}. In fact, the block partition in the BSBL framework is a \emph{regularization} for better estimation of the covariance matrix of $\mathbf{x}$ in a high-dimensional parameter space. Theoretically explaining the empirical strategy in the BSBL framework is an important topic in the future.

\subsection{Reconstruction of Non-Sparse Signals}

Most raw physiological signals are not sparse, especially when contaminated by various noise. To reconstruct these non-sparse signals, there are two popular strategies.

One is using thresholding \cite{Dixon2012} to set entries of small amplitudes to zero. However, these thresholding methods cannot be used for FECG recordings. As we have seen, the amplitudes of FECGs are very small and even invisible. Thus it is difficult or even impossible to choose an optimal threshold value. What's worse is that the thresholding methods can destroy interdependence structure among multichannel recordings, such as the ICA mixing structure.

Another strategy widely used by CS algorithms is reconstructing signals first in transformed domains, as expressed in (\ref{equ:CS_model}). The success of this strategy strongly depends on the sparsity level of the representation coefficients $\boldsymbol{\theta}$. Unfortunately, for most raw physiological signals, the representation coefficients $\boldsymbol{\theta}$ are still not sparse enough \cite{zhang2012TBME_EEG}; although coefficients of large amplitudes are  few, the number of coefficients of small amplitudes is very large. When reconstructed signals are going to be further processed by other signal processing/machine learning techniques, reconstructing these coefficients of small amplitudes is important. As shown in this work, the failure to reconstruct these coefficients resulted in the failure of ICA to extract FECGs.

The BSBL-BO algorithm, unlike existing algorithms, directly reconstructs non-sparse signals without resorting to the above two strategies. Its reconstruction with high quality allows further signal processing or pattern recognition for clinical diagnosis. Clearly, exploiting block structure and intra-block correlation plays a crucial role in the reconstruction.

\subsection{Energy-Saving by the BSBL Framework}
\label{subsec:energy}

This work focuses on algorithms for wireless FECG telemonitoring. It does not involve the analysis of energy consumption, such as the comparison between BSBL-BO and wavelet compression. However, this issue actually has been addressed in the work by Mamaghanian et al. \cite{mamaghanian2011compressed}. According to their  `digital CS' paradigm, if two CS algorithms use the same sensing matrix, their energy consumption is the same. Since in most experiments we used the same sparse binary matrix as theirs (12 entries of 1s in each column of $\mathbf{\Phi}$), their analysis on the energy consumption and their comparison between their CS algorithm and wavelet compression  are applicable to BSBL-BO.

But BSBL-BO can further reduce the energy consumption while maintaining the same reconstruction performance. In Section \ref{subsec:d} we have shown that BSBL-BO has the same performance regardless of the values of $d$ ($d$ is the number of entries of 1s in each column of $\mathbf{\Phi}$). Thus we can use a sparse binary sensing matrix with $d=2$ to save more energy.

For example, when compressing a signal of 512 time points to 256 time points, using a sparse binary sensing matrix with $d=2$ only needs about 768 additions, while using a sparse binary sensing matrix with $d=12$ requires about 5888 additions. Thus, using the sparse binary matrix with $d=2$ can greatly reduce code execution in CPU, thus reducing energy consumption. Note that when using a Daubechies-4 Wavelet to compress the signal, it requires 11784 multiplications and 11272 additions. In addition, the seeking of wavelet coefficients of large amplitudes also costs extra energy.

It should be noted that it seems that only BSBL-BO (and other algorithms derived from the BSBL framework) can use such a sparse binary sensing matrix with $d=2$ to compress signals. Our experiments on adult ECGs \footnote{Since the compared ten CS algorithms failed to reconstruct FECG recordings, we used adult ECGs without noise in the experiments. Due to space limit the results are omitted here.} showed that other CS algorithms failed to reconstruct or had degraded reconstruction quality when using this sensing matrix. In \cite{mamaghanian2011compressed} it is also shown that the basis pursuit algorithm was very sensitive to $d$; when $d$ decreased from 12 to 2, the reconstruction performance measured by output SNR decreased from 20 dB to 7 dB (when the sensing matrix was of the size $256\times 512$).


\subsection{Significance of the BSBL Framework}
The ability of the BSBL framework to recover non-sparse
signals has interesting mathematical implications. By linear
algebra, there are infinite solutions to the underdetermined
problem (\ref{equ:SMV model_noiseless}). When the true solution $\mathbf{x}_{\mathrm{true}}$ is sparse, using CS
algorithms it is possible to find it. But when the true solution
$\mathbf{x}_{\mathrm{true}}$ is non-sparse, finding it is more challenging and new constraints/assumptions
are called for. This
work shows that when exploiting the block structure and the intra-block
correlation of $\mathbf{x}_{\mathrm{true}}$, it is possible to find a solution
$\widehat{\mathbf{x}}$ which is very close to the true solution $\mathbf{x}_{\mathrm{true}}$. These findings raise new
and interesting possibilities for signal compression as well as theoretical questions in the subject of sparse and non-sparse
signal recovery from a small number of measurements (i.e., the compressed data $\mathbf{y}$).

\section{Conclusions}
\label{sec:conclusions}

FECG telemonitoring via wireless body-area networks with low-energy constraint is a challenge for CS algorithms. This work showed that the block sparse Bayesian learning framework can be successfully employed in this application.  Its success relies on two unique abilities; one is the ability to reconstruct non-sparse structured signals, and the other is the ability to explore and exploit correlation structure of signals to improve performance. Although the focus is the wireless FECG telemonitoring, the proposed framework and associated algorithms can be used to many other telemedicine applications, such as telemonitoring of adult ECGs \cite{mamaghanian2011compressed}, wireless electroencephalogram \cite{zhang2012TBME_EEG}, and electromyography \cite{Dixon2012}.

\bibliographystyle{IEEEtran}

\bibliography{spatio,ecg,sparse,zhilin}

\end{document}